\documentclass[10pt,twocolumn,letterpaper]{article}
\usepackage{cvpr}              
\usepackage[accsupp]{axessibility} 
\definecolor{cvprblue}{rgb}{0.21,0.49,0.74}
\usepackage[pagebackref,breaklinks,colorlinks,allcolors=cvprblue]{hyperref}
\usepackage{graphicx}
\usepackage{amsmath}
\usepackage{amssymb}
\usepackage{booktabs}
\usepackage{multirow}
\usepackage{algorithmic}
\usepackage[ruled]{algorithm2e} 

\def\paperID{14767} 
\def\confName{CVPR}
\def\confYear{2025}

\newcommand{\figref}[1]{Fig.~\ref{#1}}
\newcommand{\tabref}[1]{Tab.~\ref{#1}}
\newcommand{\secref}[1]{Sec.~\ref{#1}}
\newcommand{\algref}[1]{Alg.~\ref{#1}}

\newcommand*{\affaddr}[1]{#1} 

\newcommand*\samethanks[1][\value{footnote}]{\footnotemark[#1]}

\usepackage[symbol]{footmisc}
\usepackage[accsupp]{axessibility} 
\title{Style-Editor: Text-driven object-centric style editing}

\author{
\large
Jihun Park\footnote{}, Jongmin Gim\samethanks~, Kyoungmin Lee\samethanks, Seunghun Lee and Sunghoon Im\footnote{}\\
\normalsize{\affaddr{DGIST, Daegu, Republic of Korea}}\\
{\tt\small \{pjh2857, jongmin4422, kyoungmin, lsh5688, sunghoonim\}@dgist.ac.kr}
}

\begin{document}
\twocolumn[{
\maketitle
\begin{center}
    \captionsetup{type=figure}
        \vspace{-0.2cm}
    \includegraphics[width=.95\linewidth]{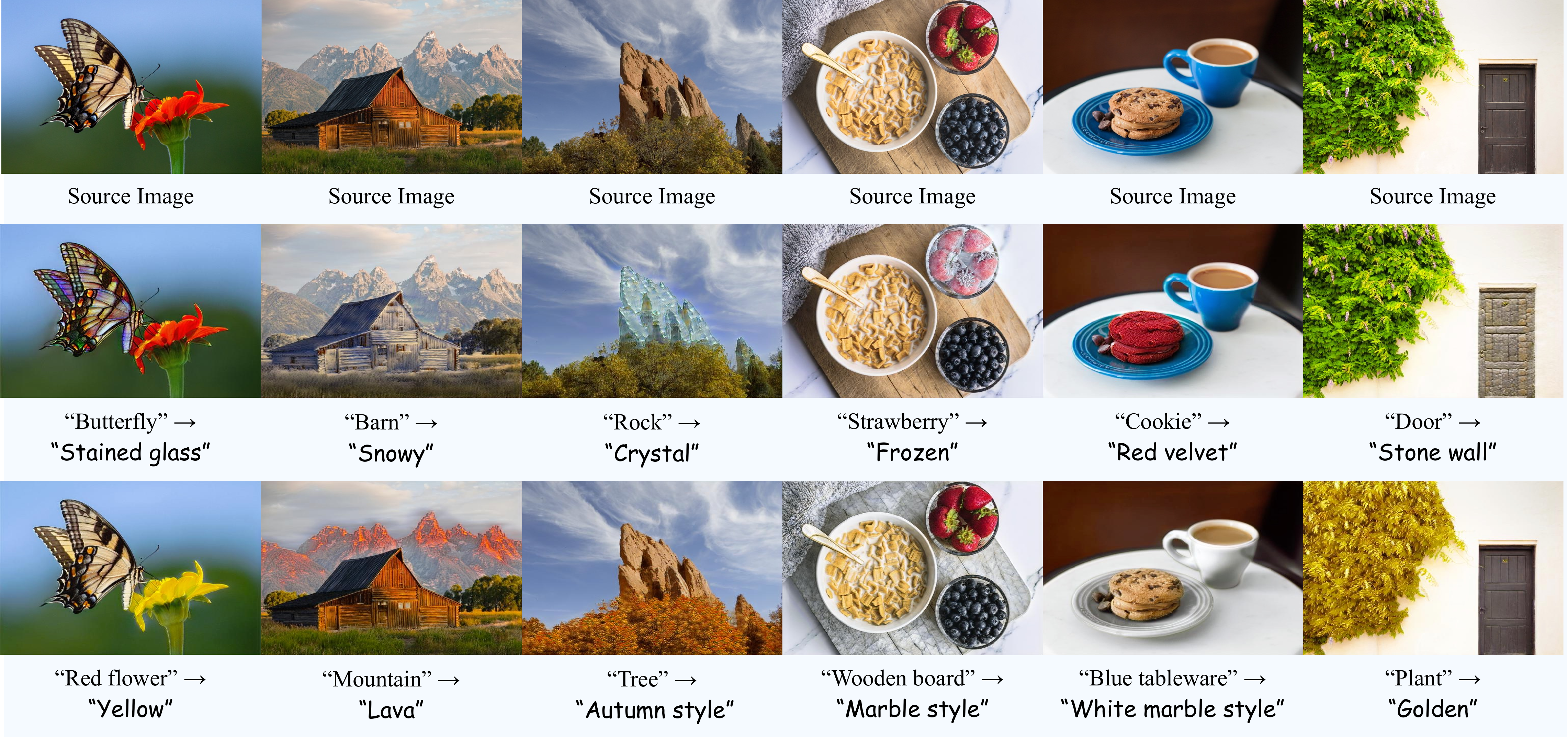}
    \captionof{figure}{Results of our Style-Editor under diverse textual conditions. To the left and right of the arrow($\rightarrow$) indicate the source text $T^{\text{src}}$ and style text $T^{\text{sty}}$, respectively. }
    \label{fig:teaser}
\end{center}
}]

\begin{abstract}
We present Text-driven object-centric style editing model named Style-Editor, a novel method that guides style editing at an object-centric level using textual inputs.
The core of Style-Editor is our Patch-wise Co-Directional (PCD) loss, meticulously designed for precise object-centric editing that are closely aligned with the input text. This loss combines a patch directional loss for text-guided style direction and a patch distribution consistency loss for even CLIP embedding distribution across object regions. 
It ensures a seamless and harmonious style editing across object regions.
Key to our method are the Text-Matched Patch Selection (TMPS) and Pre-fixed Region Selection (PRS) modules for identifying object locations via text, eliminating the need for segmentation masks. 
Lastly, we introduce an Adaptive Background Preservation (ABP) loss to maintain the original style and structural essence of the image’s background. This loss is applied to dynamically identified background areas.
Extensive experiments underline the effectiveness of our approach in creating visually coherent and textually aligned style editing.
\end{abstract}

\footnotetext[1]{Equal contribution.}
\footnotetext[2]{Corresponding author.}

\renewcommand*{\thefootnote}{\arabic{footnote}}    
\section{Introduction}
\label{sec:intro}

\begin{figure}[t]
    \includegraphics[width=1\linewidth]{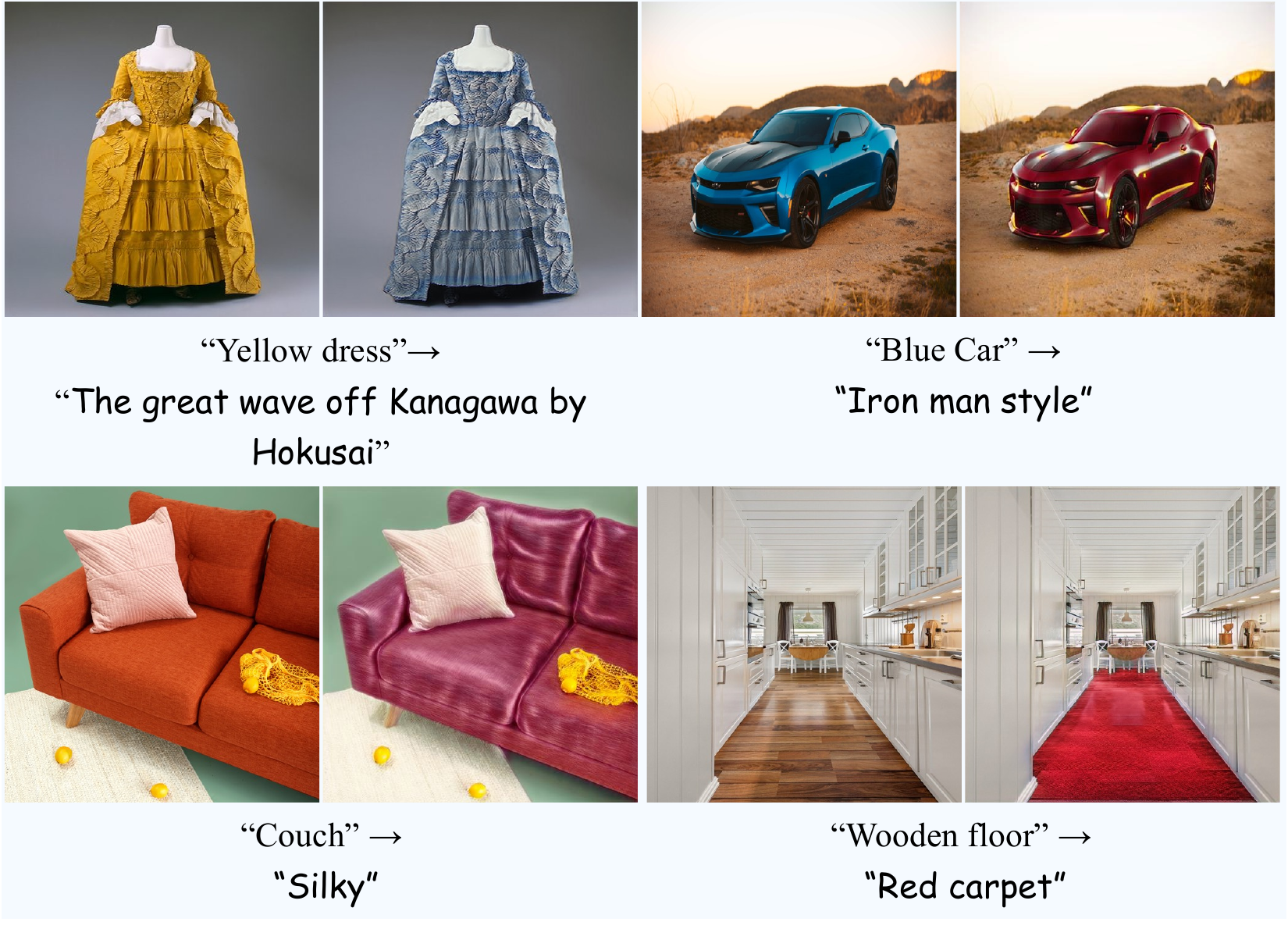}\\
    \vspace{-4mm}
    \captionof{figure}{Our editing results in industrial applications. }
    \label{fig:industry}
    \vspace{-4mm}
\end{figure}

In the realm of creative digital industries such as advertising, film, and video game development, the demand for advanced image manipulation is surging. The introduction of an object-focused style editing model, driven by textual commands, is transforming these sectors. It allows for detailed and user-friendly adjustments to the visual aspects of objects in images. This innovation empowers designers to bypass traditional manual editing, enabling them to define stylistic alterations using just text, thus facilitating rapid concept development and customization. Consider the ease with which digital fashion elements can be modified, car hues altered for online showrooms, or furniture designs changed in virtual environments, all through simple text instructions, as illustrated in \figref{fig:industry}. 

Style editing began with the foundational concept of style transfer, which traditionally relies on reference images to guide the transformation process, as demonstrated by seminal works \cite{Gatys_2016_CVPR, johnson2016perceptual, ulyanov2016texture, dumoulin2016learned, ulyanov2016instance, li2017diversified, chen2021artistic}. While these methods have been foundational, their dependency on visual templates can constrain creative possibilities. In contrast, a burgeoning wave of research in text-guided style transfer \cite{kamra2023sem, kwon2022clipstyler, bar2022text2live, fu2021language, jandial2023gatha, yu2022towards} is redefining the landscape. By dispensing with the need for reference images, these novel approaches broaden the horizons for stylistic editing, harnessing the power of textual descriptions to steer the creative process. The pioneering approach \cite{bar2022text2live, brooks2023instructpix2pix,parmar2023zero, tumanyan2023plug, mokady2023null} empowers object-centric editing directly steered by textual descriptions. Yet, this method also presents certain drawbacks; it risks altering the original content's fidelity, may fail to accurately capture the intended textual descriptions, or may unintentionally edit undesired parts of the image.

To address this issue, we introduce Text-driven object-centric style editing model named Style-Editor, a novel approach aimed at transforming the appearance of objects based on textual descriptions. This approach is designed to retain the structural integrity of objects and preserve both the appearance and structure of the background.
At the core of Style-Editor lie three pivotal components: the Patch-wise Co-Directional (PCD) loss, the Adaptive Background Preservation (ABP) loss, and the Text-Matched Patch Selection (TMPS) with Pre-fixed Region Selection (PRS) module.

Leveraging the robust zero-shot image classification capabilities of the CLIP model,
the TMPS and PRS modules are designed to identify and style the locations of objects related to the text. Unlike traditional directional loss \cite{gal2022stylegan, kwon2022clipstyler, yang2023zero} that may cause distortion due to a focus on vector directions, our PCD loss, augmented by the TMPS module, directs style editing in foreground regions aligned with the source text. This ensures a consistent CLIP-embedding distribution across patches from both source and stylized images, promoting a cohesive style editing within object areas.
Our ABP loss is designed to preserve the style and structure of the background in the source image, specifically in non-object areas. It adaptively targets and applies the loss to dynamically detected background regions, ensuring a natural and seamless transition in object-centric style editing and blending stylized elements with their surroundings as shown in \figref{fig:industry}.

In summary, our primary contributions include:
\begin{itemize}
    \item We present the Text-Matched Patch Selection (TMPS) and Pre-fixed Region Selection (PRS), which identify regions of objects related to text for object-centric style editing.
    \item We propose the Patch-wise Co-Directional (PCD) loss to enable precise style editing on targeted objects, maintaining the integrity and coherence of the source's visual aesthetic.
    \item We introduce the Adaptive Background Preservation (ABP) loss, effectively maintaining the original style and structure of designated background areas.
\end{itemize}

\section{Related Works}

\subsection{Style Transfer} 
Neural Style Transfer (NST) represents a significant advancement in the domain of image stylization, introduced by \cite{Gatys_2016_CVPR}. This pioneering approach utilized a pre-trained Convolutional Neural Network (CNN), particularly VGGNet, to extract distinct content and style features. However, a notable constraint of NST lies in its substantial computational demand, stemming from the per-image optimization approach. To overcome this, \cite{Huang_2017_ICCV} introduced Adaptive Instance Normalization (AdaIN), a technique that aligns the mean and variance of source image features with those of the style image. Building on this, \cite{li2017universal, li2018closed} proposed the Whitening and Coloring Transform (WCT), which aligns the entire covariance matrix of the features, resulting in more refined and superior stylization outcomes. With the advent of attention mechanisms in neural networks \cite{vaswani2017attention, dosovitskiy2020image}, new style transfer models have emerged that utilize these mechanisms to achieve impressive results \cite{liu2021adaattn, hong2023aespa, deng2020arbitrary, park2019arbitrary, yao2019attention}, reflecting the ongoing evolution of style transfer technology.

\begin{figure*}[t]
    \centering
    \includegraphics[width=1\textwidth]{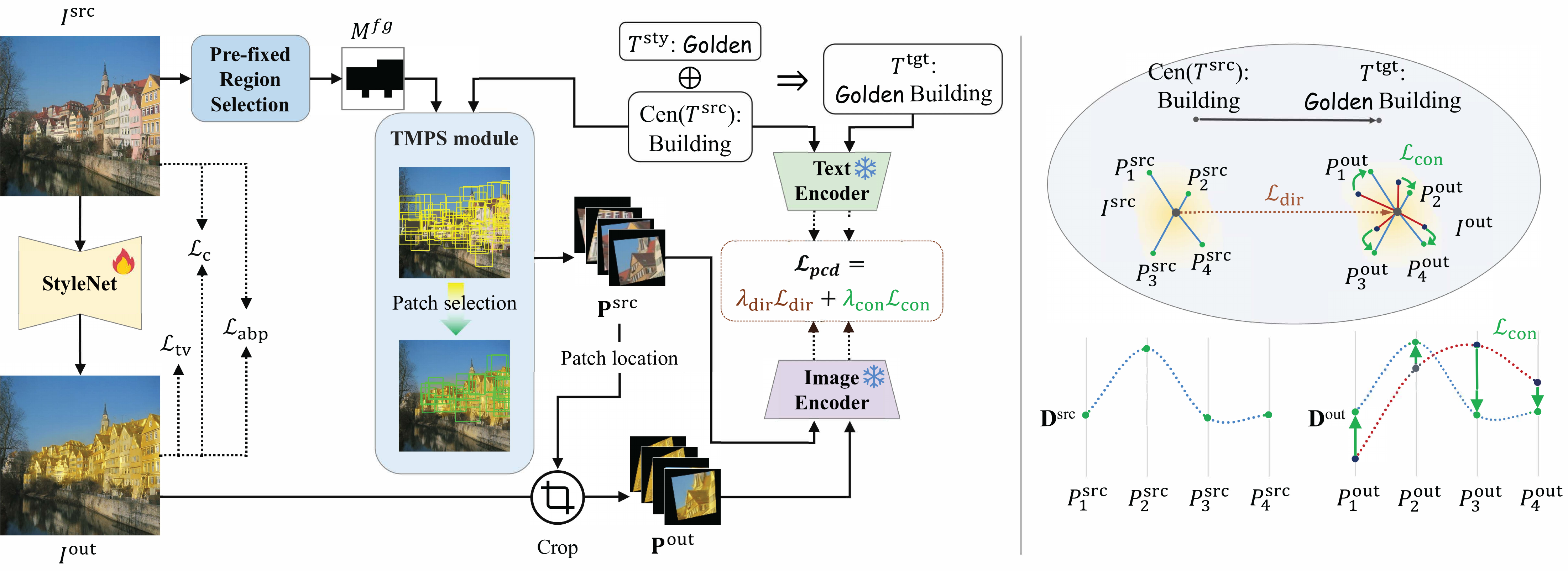}
    \captionof{figure}{\textbf{(Left)} Overall pipeline of our Style-Editor consisting of a style editing network (StyleNet), Pre-fixed Region Selection (PRS), Text-Matched Patch Selection (TMPS) module and pretrained CLIP encoders.
    The StyleNet takes a source image \(I^{\text{src}}\) and generates an object-wise stylized image \(I^{\text{out}}\).
    The TMPS module is responsible for pinpointing patches that most closely correspond to \(T^{\text{src}}\) from the foreground regions identified by the PRS.
    The selected and augmented patches, \(\mathbf{P}^{\text{src}}, \mathbf{P}^{\text{out}}\), are then aligned with \(T^{\text{src}}, T^{\text{tgt}}\) in the CLIP embedding space using Patch-wise Co-Directional (PCD) loss \(\mathcal{L}_{\text{pcd}}\). The target text \(T^{\text{tgt}}\) is derived by central word selection.
    Additionally, we apply a content loss \(\mathcal{L}_{\text{c}}\), an Adaptive Background Preservation (ABP) loss \(\mathcal{L}_{\text{abp}}\) to enhance object-centric style editing, along with a total variance loss \(\mathcal{L}_{\text{tv}}\) for regularization. 
    \textbf{(Right)} Illustration of the functionality of the PCD loss \(\mathcal{L}_{\text{pcd}}\) in feature space. It is composed of a patch-wise directional loss \(\mathcal{L}_{\text{dir}}\) and a patch distribution consistency loss \(\mathcal{L}_{\text{con}}\).
    }
    \label{fig:model_arch}
\end{figure*}

\subsection{Text-Guided Image Synthesis}
The Contrastive Language-Image Pretraining (CLIP) model, trained on a large-scale image-text dataset \cite{radford2021learning}, has significantly impacted image synthesis applications \cite{pinkney2022clip2latent, schaldenbrand2022styleclipdraw, guo2023zero, frans2022clipdraw}. Research \cite{patashnik2021styleclip, gal2022stylegan} building upon the StyleGAN architecture \cite{karras2019style, karras2020analyzing} has shown how text descriptors can adapt the style of source images. For instance, the directional loss introduced in \cite{gal2022stylegan} showcases remarkable stability and diversity in this domain, overcoming the previous requirement of a style image for transfer.
Parallel to these developments, the field has seen a surge in research focusing on diffusion model \cite{ho2020denoising}.

Recent works \cite{nichol2021glide, yang2023zero, saharia2022photorealistic,ruiz2023dreambooth,couairon2022flexit} are exploring new frontiers in image editing and generation, using text-driven diffusion models.
Despite these advancements, a persisting challenge for both Generative Adversarial Networks (GANs) and diffusion models is maintaining the subject's identity consistently in the generated images.

\subsection{Text-driven Object-Centric Style Editing}

Style editing modifies the color or texture of the original image to match a desired style while preserving the image's content \cite{huang2024diffusion}. 
Early methods \cite{kurzman2019class, huo2021manifold, huang2019style} utilized reference images for style information, applying style editing to targeted areas through segmentation networks or clustering methods.
StyleGAN-NADA \cite{gal2022stylegan} introduced a paradigm shift by eliminating the need for reference images and instead leveraging textual directions in the CLIP space for domain transfer.
Building on this, CLIPstyler \cite{kwon2022clipstyler} successfully applied this concept in text-driven style editing, advancing this field.
Following this trajectory, \cite{bar2022text2live} proposed a text-driven object-centric style editing approach by extracting a relevancy map \cite{chefer2021generic}. 

Another approach to text-driven object-centric style editing involves diffusion models trained on large vision-text datasets, such as \cite{rombach2022high, saharia2022photorealistic}. Within these approaches, some involve training the model on newly generated text-image pairs \cite{brooks2023instructpix2pix} or optimizing null-text \cite{mokady2023null}. Additionally, methods \cite{tumanyan2023plug, parmar2023zero, brack2024ledits++, patashnik2023localizing} have been developed to edit feature maps of the diffusion model, enabling more precise control over style editing based on specific textual guidance. Despite these diverse advancements, a notable limitation remains: these techniques often lead to unintended alterations of both style and content, and they often fail to achieve high levels of style editing fidelity.

\section{Method}
\subsection{Overview Framework}

The overall pipeline of our model is illustrated in \figref{fig:model_arch}-(Left).
We train a style editing network (StyleNet) to generate a stylized image $I^{\text{out}}$, given a source image $I^{\text{src}}$, along with accompanying source text $T^{\text{src}}$, and a style text $T^{\text{sty}}$. 
We use the text encoder and the image encoder derived from the pre-trained CLIP model, freezing their parameters during the training process. 
The training of StyleNet incorporates a composite loss function consisting of four distinct components: Patch-wise Co-Directional loss (\(\mathcal{L}_{\text{pcd}}\)), Adaptive Background Preservation loss ($\mathcal{L}_{\text{abp}}$), a content loss (\(\mathcal{L}_{\text{c}}\)), and a total variation regularization loss (\(\mathcal{L}_{\text{tv}}\)). These are weighted by the balance terms \(\lambda_{\text{abp}}, ~\lambda_{\text{c}}\) and \(\lambda_{\text{tv}}\) as follows:
\begin{equation}
\mathcal{L}_{\text{total}} =  \mathcal{L}_{\text{pcd}}
+\lambda_{\text{abp}}\mathcal{L}_{\text{abp}} +\lambda_{\text{c}} \mathcal{L}_{\text{c}} + \lambda_{\text{tv}} \mathcal{L}_{\text{tv}}.
\end{equation}
This paper primarily delves into the Patch-wise Co-Directional loss \(\mathcal{L}_{\text{pcd}}\) in \secref{sec:obj}, and the Adaptive Background Preservation loss $\mathcal{L}_{\text{abp}}$ in \secref{sec:bg}. 
The content loss $\mathcal{L}_{\text{c}}$ imposes the mean-square error between the features of an input and a stylized image extracted from the pre-trained VGG-19 networks \cite{Gatys_2016_CVPR}.
The total variation loss $\mathcal{L}_{\text{tv}}$ is a loss function designed based on the noise removal method \cite{rudin1992nonlinear}. 

\begin{algorithm}[t]
\caption{Text-Matched Patch Selection}
\label{alg:Text-matched patch selector}
\textbf{Input}: Image patch set $\mathbf{P}$ and source text \( T^{\text{src}} \) \\
\textbf{Output}: Patches corresponding to the source text \( \mathbf{P}^{\text{sel}}\) \\
\textbf{Parameter}: \( K \): \# of patches in $\mathbf{P}$, \\ \( M \): \# of patches similar to source text

\begin{algorithmic}[1]

\STATE \( \mathbf{S}, \hat{\mathbf{S}} \leftarrow \emptyset, \emptyset \)

\FOR{\(i = 1\) to \(K\)}
    \STATE \(f_i \leftarrow E_I(P_i)\) 
    \STATE \(s_i \leftarrow \frac{f_i \cdot E_T(T^{\text{src}})}{\|f_i\| \cdot \|E_T(T^{\text{src}})\|}\)
    \STATE \(\mathbf{S} \gets \mathbf{S} \cup \{s_i\}\)
\ENDFOR

\STATE \(\mathbf{I} \leftarrow \{i \mid s_i \geq \text{the } M^{\text{th}} \text{ largest value in } \mathbf{S}\}\)

\STATE \(f_{\text{avg}} \leftarrow \frac{1}{\vert \mathbf{I} \vert} \sum_{i \in \mathbf{I}} f_i\)

\FOR{\(j = 1\) to \(K\)}
    \STATE \(\hat{s}_j \leftarrow \frac{f_j \cdot f_{\text{avg}}}{\|f_j\| \cdot \|f_{\text{avg}}\|}\)
    \STATE \(\hat{\mathbf{S}} \gets \hat{\mathbf{S}} \cup \{\hat{s}_j\}\)
\ENDFOR
\STATE \(\hat{s}_k \leftarrow \text{the } (\text{round}({\frac{K}{2}}))^{\text{th}} \text{ largest value from } \hat{\mathbf{S}}\)
\STATE \({\mathbf{J}} \leftarrow \{j \mid \hat{s}_j \geq \hat{s}_k\) and \(\hat{s}_j > 0.8\}\)

\STATE $\mathbf{return}$ \( \mathbf{P}^{\text{sel}} \leftarrow \{P_{j} \mid j \in \mathbf{J}\}\)

\end{algorithmic}
\end{algorithm}

\subsection{Text-Matched Patch Selection (TMPS)}
\label{sec:fb}
A critical aspect of object-centric style editing is the accurate identification of the object's location within an image. To address this challenge, we introduce the TMPS module. This module pinpoints and selects image patches that correlate to a specified object as described by the source text, leveraging the zero-shot image classification capability of the CLIP model. Utilizing the image encoder $E_I$ and the text encoder $E_T$ from the CLIP architecture, TMPS establishes a strong link between the object and its textual descriptor. We define a representative feature vector $f_{\text{avg}}$ in \algref{alg:Text-matched patch selector}, capturing the essential characteristics of the object in the source image. Then, we identify patches $\mathbf{P}^{\text{sel}}$ with features similar to this representative vector. The detailed mechanism of TMPS is explained in \algref{alg:Text-matched patch selector}.

To streamline the search process for TMPS during the object-centric style editing, we introduce the Pre-fixed Region Selection (PRS) module within the source image. Applied during the initial iterations, this module delineates a foreground region \( M^{\text{fg}} \), configured to coarsely isolate object areas as outlined in \algref{alg:Pre-fixed Patch Selection}. 
This early demarcation of the object's location allows for generating patches specifically within the foreground region (\(M^{\text{fg}}\)) in later iterations. 
This strategy enhances the precision and efficiency of the style editing, focusing the modification on the most relevant sections of the image.

\begin{algorithm}[t]
\caption{Pre-fixed Region Selection}
\label{alg:Pre-fixed Patch Selection}
\textbf{Input}: Source image \( I^{\text{src}} \) and source text \( T^{\text{src}} \)\\
\textbf{Output}: A binary mask containing objects corresponding to the source text \( M^{\text{fg}}\) \\
\textbf{Parameter}: $\tau$: Threshold for \# of selections. 

\begin{algorithmic}[1]
\STATE Divide \(I^{\text{src}}\) into \(L\) uniform square grids.
\STATE Generate a set of three distinct-sized patches per grid: \\ \( \mathbf{P}^{\text{grid}} = \{P^{\text{grid}}_1, \ldots, P^{\text{grid}}_{3L}\} \).
\STATE Obtain selected patches \(\mathbf{P}^{\text{grid\_sel}}\) using TMPS module: \\ \(\mathbf{P}^{\text{grid\_sel}} = \textit{TMPS}(\mathbf{P}^{\text{grid}}, T^{\text{src}})\).
\STATE Initialize a voting matrix \(V \in \mathrm{R}^{H\times W}\) with all elements set to zero.
\FOR{each pixel in the selected patches \(\mathbf{P}^{\text{grid\_sel}}\)}
    \STATE Increment the corresponding element in \(V\).
\ENDFOR
\STATE Determine the pre-fixed foreground region \(M^{\text{fg}}\):
\[ M^{\text{fg}}(i, j) = \begin{cases} 
    1, & \text{if } V(i, j) \geq \tau \\
    0, & \text{otherwise}
\end{cases} \]
\STATE $\mathbf{return}$ \( M^{\text{fg}} \)
\end{algorithmic}
\end{algorithm}

\subsection{Patch-Wise Co-Directional Loss (PCD)}
\label{sec:obj}
Our PCD loss $\mathcal{L}_{\text{pcd}}$ incorporates
a patch-wise directional loss $\mathcal{L}_{\text{dir}}$ and a patch distribution consistency loss $\mathcal{L}_{\text{con}}$ with balance terms $\lambda_{\text{dir}}$ and  $\lambda_{\text{con}}$ as follows:
\begin{equation}
    \begin{gathered}
    \mathcal{L}_{\text{pcd}} = \lambda_{\text{dir}}\mathcal{L}_{\text{dir}}+ \lambda_{\text{con}}\mathcal{L}_{\text{con}}.
    \end{gathered}
\end{equation}

\noindent \textbf{Patch-wise directional loss} The foundational concept of directional loss, as initially introduced in \cite{gal2022stylegan} and \cite{kwon2022clipstyler}, is further refined in our approach. We adapt and advance this concept specifically for object-centric style editing, with a targeted application on image patches that correspond to the source text $T^{\text{src}}$.
We commence by randomly selecting a subset of patches within the foreground region $M^{\text{fg}}$ of the source image. The selected patches, represented as $\mathbf{P}^{\text{src}} \in \{P^{\text{src}}_1,..., P^{\text{src}}_N\}$, are then extracted using our TMPS, as detailed in \algref{alg:Text-matched patch selector}.
Given the patches and the texts, the patch-wise directional loss $\mathcal{L}_{\text{dir}}$ is defined as follows:
\begin{equation}
\label{eq:dir}
    \begin{gathered}
    \mathcal{L}_{\text{dir}} = \frac{1}{N} \sum_{i=1}^N \left(1-\frac{\Delta P_i \cdot \Delta T}{|\Delta P_{i}||\Delta T|}\right), \\ 
    \Delta P_i =E_I\left(\text{aug}\left(P^{\text{out}}_i\right)\right)-E_I\left(\text{aug}\left(P^{\text{src}}_i\right)\right),\\
    \Delta T = E_T\left(T^{\text{tgt}}\right) - E_T\left(T^{\text{src}}\right),~
    T^{\text{tgt}} = T^{\text{sty}} \oplus \text{Cen}(T^{\text{src}}),
    \end{gathered}
\end{equation}
where the operator $\oplus$ denotes text combination. We utilize the function $\text{Cen}(\cdot)$ for central word selection and $\text{aug}(\cdot)$ for patch augmentation. The patches $P^{\text{out}}_i$ are derived from output stylized images $I^{\text{out}}$ using the TMPS algorithm in~\algref{alg:Text-matched patch selector}.

\noindent \textbf{Target text generation} To formulate the target text $T^{\text{tgt}}$, we apply central word selection technique $\text{Cen}(\cdot)$ in \eqref{eq:dir} leveraging Spacy \cite{honnibal-johnson-2015-improved}, and merge the source text $T^{\text{src}}$ and the style text $T^{\text{sty}}$. 
This design is rooted in the manifold augmentation technique proposed in \cite{ye2023textmania} leveraging text embeddings from models like BERT \cite{devlin2019bert}, GPT \cite{radford2018improving}, and CLIP.
This amalgamation of texts aligns more closely with the user's intended styling objectives.
For instance, to adapt the style of `red apple' to appear green, we engage TMPS using `red apple' but modify the target text to `green apple', with an emphasis on `apple' as the central word. 
This refined approach to text input in TMPS enables more accurate patch selection, facilitating precise object-centric style editing.

\noindent \textbf{Patch distribution consistency loss} Traditional directional loss functions typically emphasize the direction of vectors while often neglecting their semantic information. This emphasis can inadvertently lead to a misalignment between patches in the source image and those in the stylized image as depicted in \figref{fig:model_arch}-(Right). Such a discrepancy can lead to the collapse of semantic information, resulting in a distorted transformation that causes the loss of information from the source image, contradicting the fundamental aim of style editing. To address this issue, we design the patch distribution consistency loss $\mathcal{L}_{\text{con}}$ as follows:

\begin{equation}
    \begin{gathered}
    \mathcal{L}_{\text{con}} = \text{JSD}(\mathbf{D}^{\text{src}},\mathbf{D}^{\text{out}}), \\
    \mathbf{D}^{\text{src}}=\frac{\hat{\mathbf{D}}^{\text{src}}}{\sum_{i=1}^{N} \hat{\mathbf{D}}_i^{\text{src}}},~ \hat{\mathbf{D}}^{\text{src}}=\left(\frac{E_I\left(P^{\text{src}}_i\right)\cdot E_I(I^{\text{src}})}{|E_I\left(P^{\text{src}}_i\right)|\cdot |E_I(I^{\text{src}})|}\right)_{i=1}^N, \\ \mathbf{D}^{\text{out}}=\frac{\hat{\mathbf{D}}^{\text{out}}}{\sum_{i=1}^{N} \hat{\mathbf{D}}_i^{\text{out}}}, ~    \hat{\mathbf{D}}^{\text{out}}=\left(\frac{E_I\left(P^{\text{out}}_i\right)\cdot E_I(I^{\text{out}})}{|E_I\left(P^{\text{out}}_i\right)|\cdot |E_I(I^{\text{out}})|}\right)_{i=1}^N,
    \end{gathered}
\end{equation}
where $\text{JSD}(\cdot)$ is Jensen–Shannon divergence, which is employed to align the feature distributions of patches $\mathbf{D}^{\text{src}}$ from the source image with those $\mathbf{D}^{\text{out}}$ from the stylized image.

\begin{table*}[t]
\centering
{
\begin{tabular}
{c@{\hspace{1mm}}|c@{\hspace{5mm}}c@{\hspace{5mm}}|c@{\hspace{5mm}}c@{\hspace{5mm}}c@{\hspace{5mm}}c@{\hspace{5mm}}c@{\hspace{5mm}}c@{\hspace{3mm}}}
\toprule
   & \multicolumn{2}{c|}{Foreground quality metrics} &\multicolumn{6}{c}{Background quality metrics}\\
  Methods  & \multicolumn{1}{c}{\hspace{4mm} $\text{Sim}_{F}\uparrow$} & \multicolumn{1}{c|}{$\text{Con}_{F}\downarrow$} & \multicolumn{1}{c}{$\text{L1}_{B}\downarrow$} & \multicolumn{1}{c}{$\text{Con}_{B}\downarrow$} & \multicolumn{1}{c}{$\text{Sty}_{B}\downarrow$} & \multicolumn{1}{c}{$\text{SSIM}_{B}\uparrow$} & \multicolumn{1}{c}{$\text{DISTS}_{B}\downarrow$} & \multicolumn{1}{c}{$\text{PSNR}_{B}\uparrow$} \\
\midrule
  FlexIT~\cite{couairon2022flexit} & \hspace{4mm} 0.24 & 8.08 & 0.25 & 4.78 & 0.73 & 0.60 & 0.17 & 19.69 \\
  ZeCon~\cite{yang2023zero}  & \hspace{4mm} 0.23 & 7.49 & 0.28 & 5.33 & 0.63 & 0.64 & 0.24 & 19.35 \\
  Null-text inversion~\cite{mokady2023null} & \hspace{4mm} 0.20 & 4.22 & 0.16 & 2.58 & 0.22 & 0.74 & 0.10 & 23.48 \\
  Instruct Pix2Pix~\cite{brooks2023instructpix2pix} & \hspace{4mm} 0.22 & 7.42 & 0.44 & 4.66 & 0.97 & 0.62 & 0.20 & 17.25 \\
  Plug and Play~\cite{tumanyan2023plug} & \hspace{4mm} 0.23 & 6.51 & 0.33 & 4.26 & 0.63 & 0.63 & 0.22 & 18.26 \\
  pix2pix-zero~\cite{parmar2023zero} & \hspace{4mm} 0.22 & 8.12 & 0.40 &  4.91 &  1.07 &  0.61 &  0.26 &  17.11 \\
  LEDITS++~\cite{brack2024ledits++} & \hspace{4mm} 0.22 &  6.81 & 0.18 &  2.92 &  0.44 &  0.74 &  0.14 &  21.66 \\
  local-prompt-mixing \cite{patashnik2023localizing}& \hspace{4mm} 0.20 &  9.65 & 0.24 &  4.65 &  0.61 &  0.67 &  0.18 &  20.01 \\
  \midrule
  StylerDALLE~\cite{xu2023stylerdalle} & \hspace{4mm} 0.26 & 8.35 & 0.51 & 5.71 & 1.22 & 0.52 & 0.28 & 15.42 \\
  CLIPstyler~\cite{kwon2022clipstyler}  & \hspace{4mm} 0.28 & 5.16  & 0.66 & 3.27 & 0.35 & 0.51 & 0.25 & 13.20 \\
  Text2LIVE~\cite{bar2022text2live} & \hspace{4mm} \underline{0.32} & \underline{4.13} & \underline{0.14} & \underline{1.22} & \underline{0.18} & \underline{0.87} & \underline{0.09} & \underline{24.69} \\\midrule
  Ours  & \hspace{4mm} \textbf{0.33} & \textbf{3.75} & \textbf{0.10} & \textbf{1.15} & \textbf{0.10} & \textbf{0.90} & \textbf{0.07} & \textbf{27.65} \\
\bottomrule
\end{tabular}
\caption{Quantitative comparison of our method with the recent text-guided style editing methods using foreground quality metrics and background quality metrics. The symbols $\uparrow$ and $\downarrow$ indicate higher values are better and lower values are better, respectively.}
\label{tab:vs_text}
}
\end{table*}

\subsection{Adaptive Background Preservation (ABP) Loss}
\label{sec:bg}
We introduce the ABP loss, designed to ensure the non-style editing of the background region. 
Our approach begins by identifying foreground regions as outlined in \algref{alg:Text-matched patch selector}, \algref{alg:Pre-fixed Patch Selection}.
In each iteration, the adaptive foreground mask $M^{\text{fg*}}$ and background mask $M^{\text{bg*}}$ are dynamically updated as follows:
\begin{equation}
\begin{gathered}    
M^{\text{bg*}} = 1-M^{\text{fg*}},~
M^{\text{fg*}} = \left( \bigvee_{i=1}^{N_{\text{iter}}} M^{\text{src}}_i \right),
\end{gathered}
\end{equation}
where $N_{\text{iter}}$ represents the number of patches in each iteration, and the binary mask $M^{\text{src}}_i$ is assigned a value of one over the area of patches $\{P^{\text{src}}_i\}_{i=1,...N_{\text{iter}}}$, which are selected from the source image through the TMPS.

Next, we apply the MS-SSIM and L1 loss functions to ensure that the original styles of background regions are retained as follows:
\begin{equation}
\begin{gathered}    
\mathcal{L}_{\text{abp}} = \mathcal{L}_{\text{MS\_SSIM}}(I^{\text{out}}\odot M^{\text{bg*}}, I^{\text{src}}\odot M^{\text{bg*}}) \\ 
+ \mathcal{L}_{\text{L1}}(I^{\text{out}}\odot M^{\text{bg*}}, I^{\text{src}}\odot M^{\text{bg*}}).
\end{gathered}
\end{equation}
This loss enables a well-balanced integration of the stylized and original regions within the image. Specifically, while the foreground areas undergo style editing, the background's integrity is preserved.

\begin{figure*}[t]
    \centering
    \includegraphics[width=.95\textwidth]{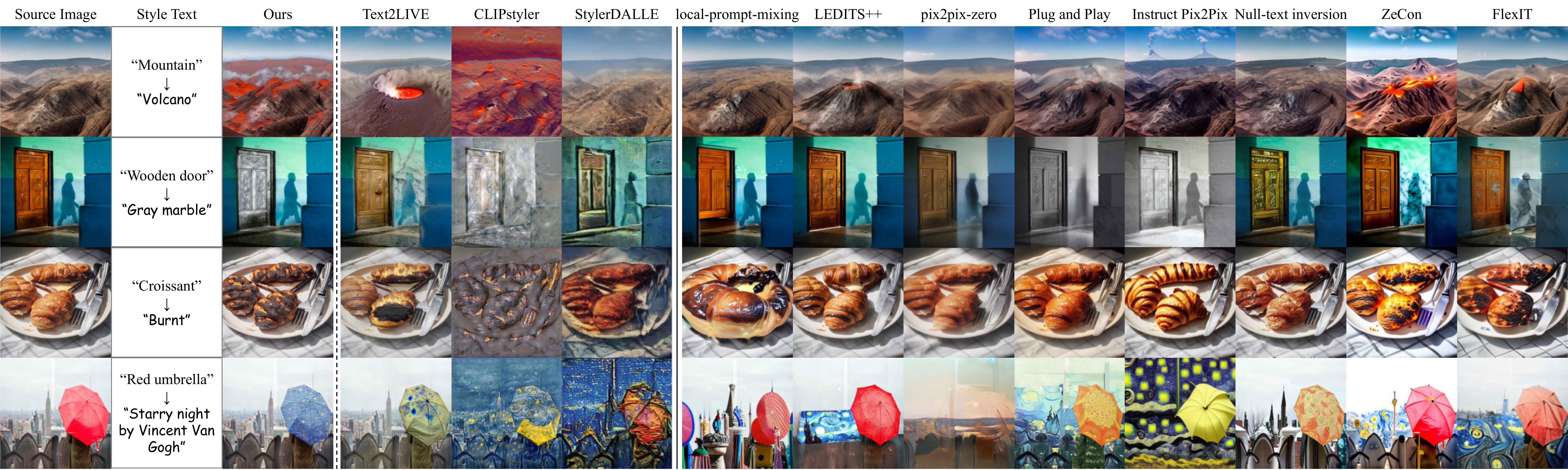} \\ 
    \captionof{figure}{Comparison of our method with various text-guided style editing models. To the left of the solid line are the qualitative results of our model and non-diffusion based models, to the right are the results from diffusion-based methods.}
    \label{fig:vs_text}
\end{figure*}

\section{Experiments}
\subsection{Implementation Details}
We employ the pretrained CLIP model on VIT-B/32 \cite{dosovitskiy2020image}. Our stylization network is based on the U-net architecture \cite{ronneberger2015u} consisting of three downsample and three upsample layers featuring channel sizes of 16, 32, and 64. The source images have a resolution of $512\times512$ pixels. We set $\lambda_{\text{dir}}$, $\lambda_{\text{con}}$, $\lambda_{\text{abp}}$, $\lambda_{\text{c}}$, and $\lambda_{\text{tv}}$ to $1.5\times10^{4}$, $3\times10^{4}$, $3\times10^{4}$, $4\times10^{2}$, and $2\times10^{-3}$, respectively. For content loss, we follow \cite{Gatys_2016_CVPR} by utilizing features from the ``conv4\_2'' and ``conv5\_2'' layers. 

We commence our training with an initial set of 20 early iterations and completed a total of 200 iterations using the Adam optimizer \cite{kingma2014adam}. The training began with an initial learning rate of $5\times10^{-4}$, halved after the first 100 iterations. We incorporate the perspective augmentation function from the PyTorch library \cite{paszke2019pytorch}. 
Notably, our method processes each source image independently, using only the source image, source text, and style text, without requiring additional training data. The average training time for each text prompt is approximately 45 seconds on an NVIDIA A6000 GPU. For further details on our training procedure, please refer to Appendix A.

\begin{table*}[ht]
\centering
{
\begin{tabular}{c@{\hspace{1mm}}|c@{\hspace{1mm}}c@{\hspace{1mm}}c@{\hspace{1mm}}|c@{\hspace{1mm}}c@{\hspace{1mm}}|c@{\hspace{1mm}}c@{\hspace{1mm}}c@{\hspace{1mm}}c@{\hspace{1mm}}c@{\hspace{1mm}}c@{\hspace{1mm}}c@{\hspace{1mm}}}
\toprule
  \multicolumn{4}{c|}{Components} & \multicolumn{2}{c|}{Foreground quality metrics}  & \multicolumn{6}{c}{Background quality metrics}\\
   
   \# & $\mathcal{L}_{\text{dir}}$ & $\mathcal{L}_{\text{con}}$ & $\mathcal{L}_{\text{abp}}$ & \multicolumn{1}{c}{\hspace{4mm} $\text{Sim}_{F}\uparrow$} & \multicolumn{1}{c|}{$\text{Con}_{F} \downarrow$} & \multicolumn{1}{c}{$\text{L1}_{B}\downarrow$} & \multicolumn{1}{c}{$\text{Con}_{B} \downarrow$} & \multicolumn{1}{c}{$\text{Sty}_{B}\downarrow$} & \multicolumn{1}{c}{$\text{SSIM}_{B}\uparrow$} & \multicolumn{1}{c}{$\text{DISTS}_{B}\downarrow$} & \multicolumn{1}{c}{$\text{PSNR}_{B}\uparrow$} \\
\midrule
  (a) & &  &  & 0.29 & 4.31 & 0.60 & 2.72 & 0.25 & 0.56 & 0.22 & 14.16 \\
  (b) & \checkmark & & & \underline{0.32} & 4.72 & 0.49 & 2.07 & 0.21 & 0.64 & 0.15 & 16.02 \\
 (c) & \checkmark & \checkmark & & \textbf{0.33} & 4.62 & 0.48 & 2.00 & 0.21 & 0.65 & 0.15 & 16.16 \\
 (d) & \checkmark & & \checkmark & \underline{0.32} & \underline{4.16} & \textbf{0.10} & \underline{1.28} & \underline{0.12} & \underline{0.89} & \underline{0.08} & \underline{27.28} \\
 \midrule
 (e) & \checkmark & \checkmark & \checkmark & \textbf{0.33} & \textbf{3.75} & \textbf{0.10} & \textbf{1.15} & \textbf{0.10} & \textbf{0.90} & \textbf{0.07} & \textbf{27.65} \\
\bottomrule
\end{tabular}}
\caption{Ablation study examining the effects of the proposed losses $\mathcal{L}_{\text{dir}}$, $\mathcal{L}_{\text{con}}$, and $\mathcal{L}_{\text{abp}}$. The quantitative results of (a)-(e) correspond to the qualitative results shown in \figref{fig:ablation}. The symbols $\uparrow$ and $\downarrow$ indicate higher values are better and lower values are better, respectively.}
\label{tab:ablation}
\end{table*}

\begin{figure*}[ht]
    \centering
    \includegraphics[width=1\textwidth]{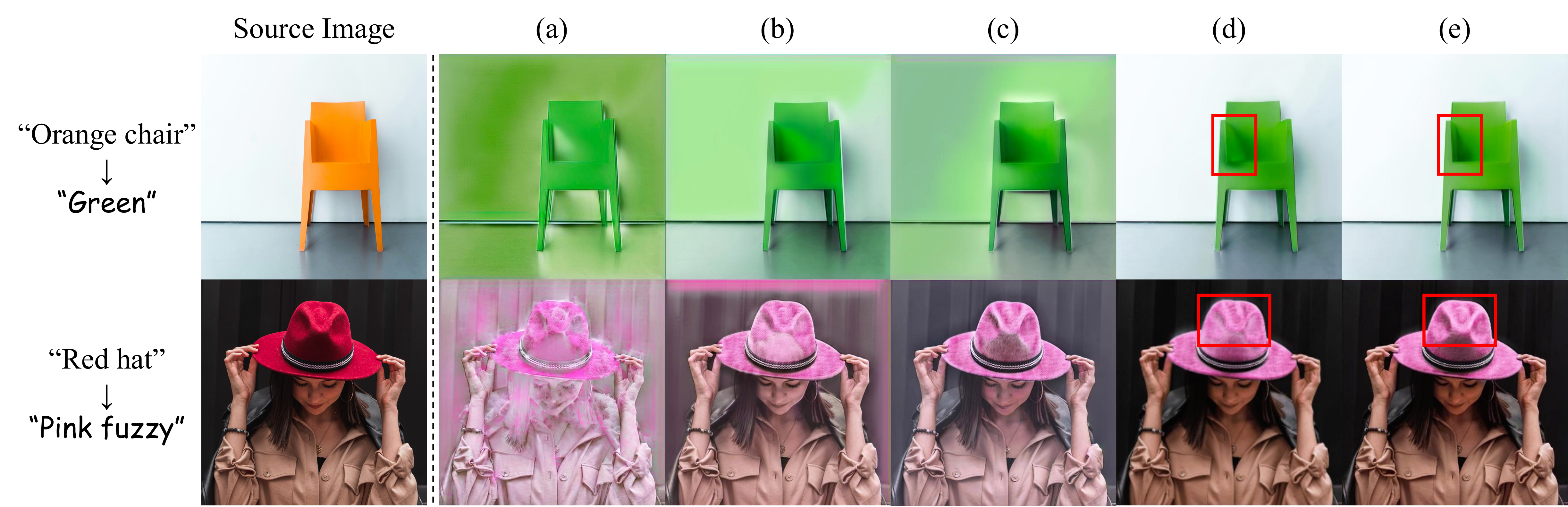} 
    \captionof{figure}{Qualitative results showcasing the impact of applying/omitting the proposed losses. Configurations (a)-(e) correspond to the settings detailed in \tabref{tab:ablation}.}
    \label{fig:ablation}
\end{figure*}

\subsection{Evaluation metrics}
We conduct a comprehensive quantitative evaluation using a variety of metrics that are well-established in the field of text-driven style editing as outlined in~\tabref{tab:vs_text}. 
 The similarity metric ($\text{Sim}$), a basic metric in the style editing field, measures the similarity between image and text embeddings.
We also apply content metrics ($\text{Con}$) using the VGG content loss \cite{johnson2016perceptual} and style metrics ($\text{Sty}$) using the mean-variance style loss \cite{huang2017arbitrary}. These metrics are adapted from traditional calculations but specifically mask either the foreground or background of input images to focus analysis on selected areas.
To thoroughly evaluate the quality of the stylized image backgrounds from multiple aspects, we conduct additional assessments specifically targeting the background masks. For this purpose, we utilize the SSIM metric \cite{zhou2004iqa}, which evaluates structural, luminance, and color attributes. Additionally, we employ the DISTS measure \cite{Ding_2020} to ensure deep structural and textural consistency in these areas. Furthermore, we use the PSNR \cite{fardo2016formal} metric and absolute difference ($\text{L1}$) for an overall assessment of signal fidelity and basic visual quality in the backgrounds.
For a comprehensive quantitative evaluation of object-centric style editing, we split the image region into foreground and background components. We utilize Ground Truth (GT) segmentation (binary) masks for the object $M^{\text{fg\_gt}}$ and background $M^{\text{bg\_gt}} (= 1 - M^{\text{fg\_gt}})$. 
For example, the $\text{Sim}_F$ and $\text{L1}_B$ are formulated as follows:
\begin{equation}
\begin{gathered}    
\text{Sim}_F = \frac{E_T\left(T^{\text{tgt}}_i\right)\cdot E_I(I^{\text{out}}\odot M^{\text{fg\_gt}})}{|E_T\left(T^{\text{tgt}}_i\right)|\cdot |E_I(I^{\text{out}}\odot M^{\text{fg\_gt}})|},\\
\text{L1}_B = \frac{\sum| I^{\text{src}}\odot M^{\text{bg\_gt}} - I^{\text{out}}\odot M^{\text{bg\_gt}} |}{255 \cdot \text{num}(I^{\text{src}}\odot M^{\text{bg\_gt}})},
\end{gathered} 
\end{equation}
where $E_{I}$ and $E_{T}$ denote the pre-trained CLIP image and text encoders, respectively. $\text{num}(\cdot)$ indicates the number of pixels in the image. 
We randomly selected 16 images from the MSCOCO 2017 dataset \cite{lin2014microsoft}, each accompanied by GT segmentation masks, to conduct evaluation. Each image was paired with a total of 10 style text descriptions based on three different categories: color (red, blue, green), texture (golden, frosted, stained glass, neon light), and artistic style (Starry Night by Vincent van Gogh, a watercolor painting of flowers, cubism style). This resulted in a total of 160 stylized images for evaluation.

\subsection{Comparison to state-of-the-arts}
In \tabref{tab:vs_text} and \figref{fig:vs_text}, we present both quantitative and qualitative comparisons of our Style-Editor model against various text-driven editing models including various diffusion models, such as  Text2LIVE \cite{bar2022text2live}, CLIPstyler \cite{kwon2022clipstyler},Null-text inversion \cite{mokady2023null}, ZeCon \cite{yang2023zero}, FlexIT \cite{couairon2022flexit}, StylerDALLE \cite{xu2023stylerdalle}, Instruct Pix2Pix \cite{brooks2023instructpix2pix}, Plug and Play \cite{tumanyan2023plug}, pix2pix-zero \cite{parmar2023zero}, LEDITS++ \cite{brack2024ledits++} and local-prompt-mixing \cite{patashnik2023localizing}. 
Our quantitative evaluation indicates that our method surpasses all compared models in performance.
Specifically, the object regions in stylized images from our method exhibit higher $\text{Sim}_F$ and lower $\text{Con}_F$ scores compared to competitive methods. This indicates that our proposed method effectively styles the object regions according to the corresponding text while preserving the structure of the source image. 
In terms of background quality, the metrics show that the proposed method subtly changes the style of background regions in the source image, as indicated by $\text{Sty}_B$. Meanwhile, it maintains the integrity of the source image's structure, as evidenced by scores from $\text{L1}_B$, $\text{Con}_B$, $\text{SSIM}_B$, $\text{DISTS}_B$, and $\text{PSNR}_B$.
This dual capability highlights our method’s adeptness at enhancing foreground elements distinctly from the background, ensuring a balanced and coherent visual output.

In the qualitative evaluation, the first row of \figref{fig:vs_text} showcases the style editing of a mountain into a volcano using the style text ``Volcano''. Our method adeptly preserves the inherent structure of the mountain while seamlessly integrating the volcanic style. In contrast, Text2LIVE often reconstructs an entirely new mountain structure, typically characterized by a singular hole indicative of magma.
This comparison underscores the nuanced capability of our approach in maintaining the original form while applying distinct style editing.
Utilizing source texts like ``wooden door'', ``croissant'' and ``red umbrella'', our method demonstrates a precise capture of object locations, contrasting with other models that often indiscriminately apply style editing to the entire image, leading to unfavorable background metric scores. Additionally, the six diffusion-based style editing models frequently alter the target object, resulting in poor $Con_{F}$ scores.
While Text2LIVE shows results similar to ours, it notably applies style editing to unintended areas (\textit{e.g.}, wooden door with ``gray marble'' as style text) and fails to properly reflect the style indicated by the style text (\textit{e.g.}, red umbrella in ''Starry night by Vincent Van Gogh'').

\subsection{Ablation study}
To demonstrate the efficacy of our method, we conduct an ablation study, with quantitative results presented in \tabref{tab:ablation}  and qualitative insights in \figref{fig:ablation}. Notably, the quantitative results in \tabref{tab:ablation}-(e) which incorporates all proposed losses $\mathcal{L}_{\text{dir}}$, $\mathcal{L}_{\text{con}}$, and $\mathcal{L}_{\text{abp}}$, indicate that our method achieves the best performance. Notably, the absence of the adaptive background preservation loss $\mathcal{L}_{\text{abp}}$ in \tabref{tab:ablation}-(c) results in a significant drop in the background absolute difference metrics $\text{L1}_B$. Omitting the patch distribution consistency loss $\mathcal{L}_{\text{con}}$ in \tabref{tab:ablation}-(d) leads to a decrease in performance across all foreground quality metrics. These findings clearly demonstrate the critical role of each proposed loss, affirming their essential contributions to our design.

The qualitative results in \figref{fig:ablation} further illustrate this point. In \figref{fig:ablation}-(a), a scenario where patches are randomly selected and directional loss based on the text direction is applied without incorporating all proposed losses and modules, demonstrates a significant limitation. Here, the entire source image undergoes style editing, which leads to a distortion of the semantic content originally present in the image.
\figref{fig:ablation}-(b) demonstrates that applying only $\mathcal{L}_{\text{dir}}$ leads to focused style editing on the targeted object, enabled by TMPS, but also results in notable background changes and some loss or alteration of object details. In contrast, \figref{fig:ablation}-(c), where $\mathcal{L}_{\text{dir}}$ is combined with $\mathcal{L}_{\text{con}}$, effectively preserves crucial object details like the chair's shadow and shape of the hat, though background alterations remain. The comparison of \figref{fig:ablation}-(d) and (e) reveals the impact of the $\mathcal{L}_{\text{con}}$; in (d), despite achieving object-centric style editing, there is a noticeable loss in object details, particularly in cropped areas.
Meanwhile, the comparison of \figref{fig:ablation}-(c) and (e) demonstrates that with the full complement of losses, both the background remains unchanged and the object details are preserved without distortion, highlighting the effectiveness of the ABP loss.

\subsection{Comparison to mask-guided generative models}

\begin{figure}[t]
    \centering
    \includegraphics[width=.5\textwidth]{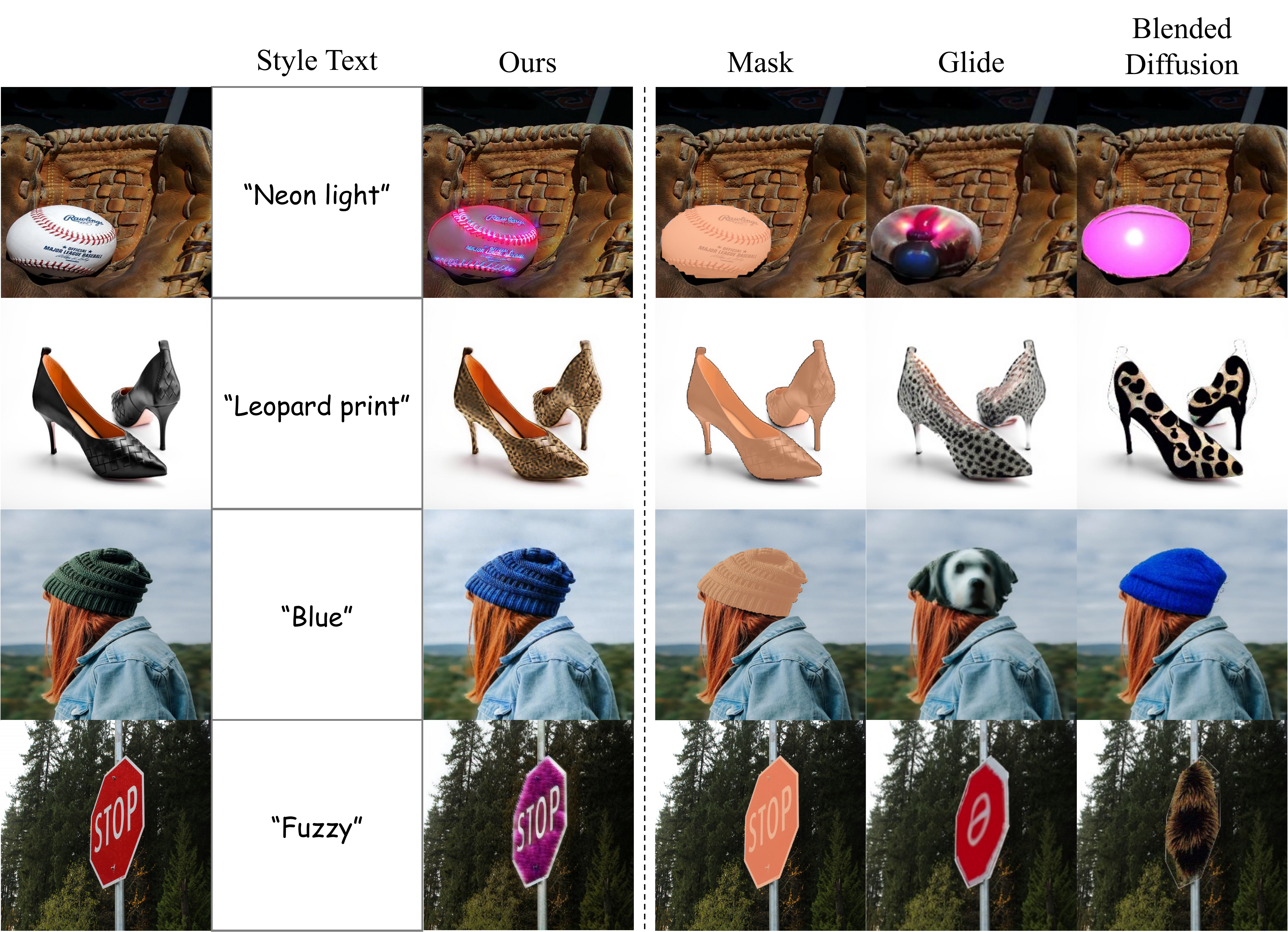} 
    \captionof{figure}{Comparative analysis of Style-Editor with other generative models using mask input. This figure demonstrates the superiority of our model in producing enhanced results compared to methods that use object masks as guidance for style editing tasks.}
    \label{fig:vs_mask_diff}
\end{figure}

To highlight the differences from mask-based models and showcase the natural style-editing capabilities of our Style-Editor model, we conduct a comparative analysis with existing mask-guided generative models \cite{avrahami2022blended, nichol2021glide}.
These models typically rely on masks to specific areas for image editing.
Our comparison, as depicted in \figref{fig:vs_mask_diff}, highlights a key distinction. Unlike mask-based models, which may inadvertently distort vital details of the object, Style-Editor consistently maintains the structural integrity of the object, focusing solely on altering its style. This process is crucial in preserving the overall coherence and authenticity of the image. Moreover, Style-Editor effectively identify relevant objects based on the provided text and apply style edits directly to these areas, eliminating the need for labor-intensive and sometimes imprecise manual mask creation. Style-Editor, thus, not only preserves the integrity of the objects but also enhances efficiency, offering a more user-friendly alternative to traditional mask-guided generative models.
\section{Conclusion}
In this paper, we have presented Text-driven Object-Centric Style editing (Style-Editor), a new approach for editing the appearance of objects in images based purely on textual descriptions. This method eliminates the need for reference style images or segmentation masks, facilitating complex and tailored style editing directly from text. At the heart of Style-Editor are three pivotal components: the Text-Matched Patch Selection with Pre-fixed Region Selection module, the Patch-wise Co-Directional (PCD) loss and the Adaptive Background Preservation (ABP) loss. The TMPS and PRS modules are pivotal in accurately identifying the location of objects linked to the source text. The PCD loss, complemented by our Text-Matched Patch Selection (TMPS), precisely identifies and selects patches for style editing, ensuring an artifact-free style editing through consistent CLIP-embedding distribution. Meanwhile, the ABP loss plays a critical role in preserving the original style and structure of the background. It adeptly adjusts to dynamically identified background areas, effectively preventing unintended alterations during the style editing process. Extensive experiments validate the superior performance of our Style-Editor, demonstrating its capability to achieve high-quality, text-driven object-centric style editing while preserving the overall visual coherence and integrity of images. 

\noindent\textbf{Acknowledgments}\\
This work was supported by the 2025 innovation base artificial intelligence data convergence project project with the funding of the 2025 government (Ministry of Science and ICT) (S2201-24-1002). The supercomputing resources for this work was supported by Grand Challenging Project of Supercomputing AI Education and Research Center, DGIST.

{
    \small
    \bibliographystyle{ieeenat_fullname}
    \bibliography{main}
}
\appendix
\onecolumn
\begin{center}
    \LARGE \textbf{Style-Editor: Text-driven object-centric style editing\\ Supplementary Material}
\end{center}

\section{Detailed training process of the Style-Editor}

\begin{figure}[h]
    \begin{minipage}{0.45\textwidth}
        \centering
        \includegraphics[width=.58\textwidth]{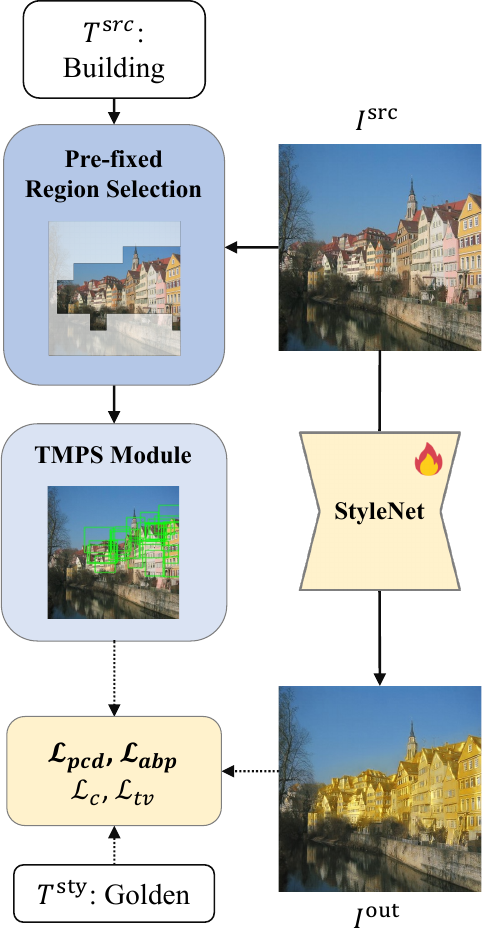}
        \captionof{figure}{Simplified overview of the training process.}
        \label{fig:pipeline}
    \end{minipage}
    \begin{minipage}{0.55\textwidth}
        \begin{itemize}
            \item The model takes three inputs: a source image ($I^{src}$), which is the subject of the style editing; a source text ($T^{src}$), which identifies the specific object in the source image to be modified; and a style text ($T^{sty}$), describing the desired style to be applied to the object.
            \item In the early iterations, the Pre-fixed Region Selection (PRS) is used to divide $I^{src}$ into coarse foreground and background regions.
            \item Within the coarsely segmented foreground region, patches are generated. The Text-Matched Patch Selection (TMPS) module then comes into play, selecting those patches that correspond to the object mentioned in $T^{src}$.
            \item The training of the model incorporates a combination of loss functions: Patch-wise Co-Directional loss ($\mathcal{L}_{\text{pcd}}$), Adaptive Background Preservation loss ($\mathcal{L}_{\text{abp}}$), Content loss ($\mathcal{L}_{\text{c}}$), and Total Variation loss ($\mathcal{L}_{\text{tv}}$). 
        \end{itemize}
    \end{minipage}
\end{figure}

\subsection{Pre-fixed Region Selection (PRS)}
\begin{figure}[h]
    \centering
    \includegraphics[width=.62\textwidth]{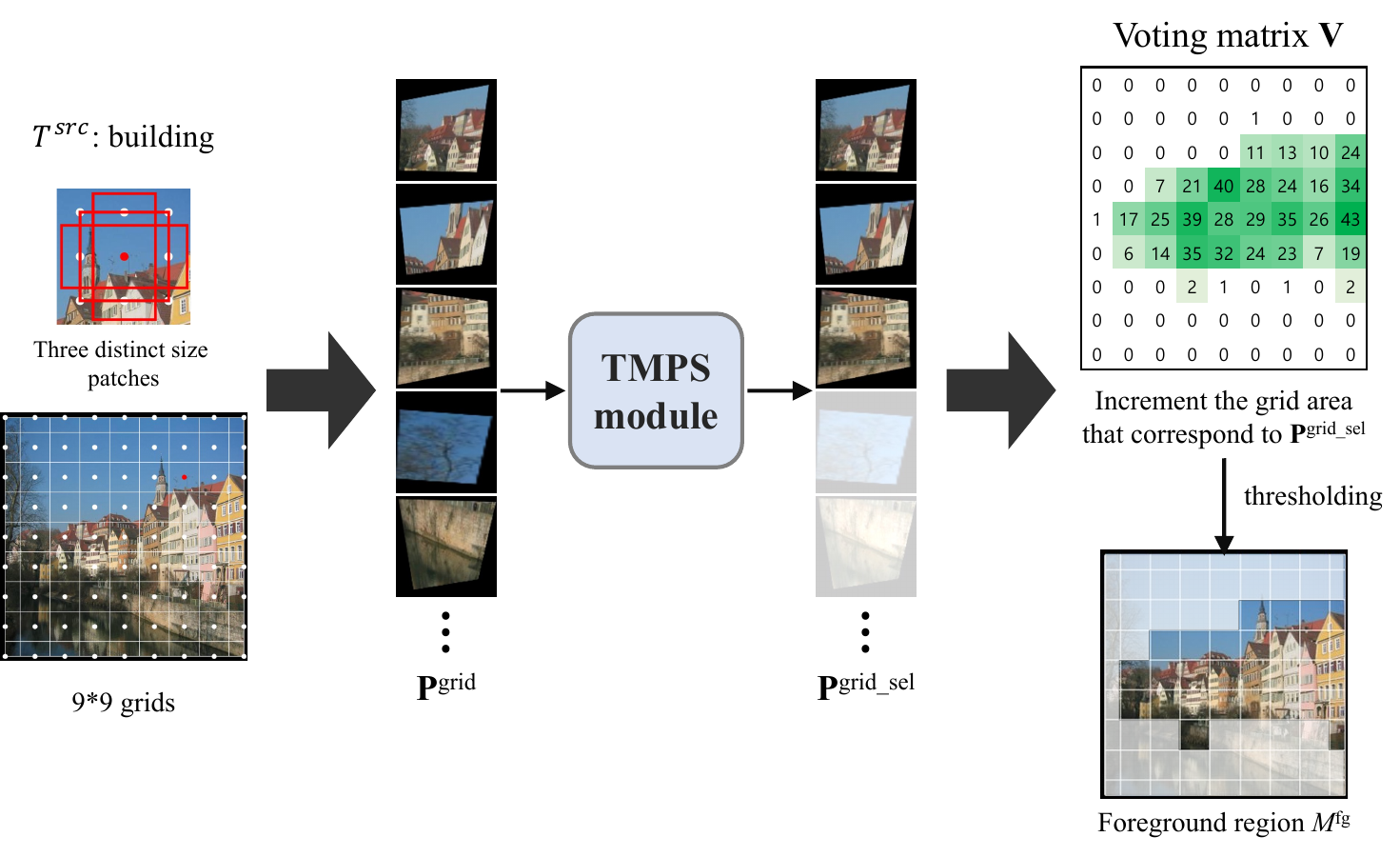} 
    \captionof{figure}{Overview of Pre-fixed Region Selection (PRS).}
    \label{fig:PRS}
\end{figure}

In our approach, we process the source image ($I^{src}$) by dividing it into a $9 \times 9$ grid. From each grid point, we generate three patches of varying sizes, centered on these points, resulting in a total of 243 patches. These patches are then passed through the Text-Matched Patch Selection (TMPS) module, which identifies and selects the patches that are most relevant to the source text ($T^{src}$).
We use a specific threshold \(\tau\) (we set as 2) to determine the foreground region ($M^{\text{fg}}$) of the image: grids that exceed this threshold in terms of patch relevance are classified as part of the foreground. After the initial iterations, further patch generation is concentrated within this coarse foreground region. The number of patches generated in subsequent iterations is adjusted proportionally to the count of grids identified as part of the foreground region. This method ensures a focused and relevant application of style editing where it is most pertinent according to the text input.

\section{Comprehensive Analysis of our Style-Editor}

In this section, we provide a detailed analysis of our Style-Editor model by evaluating its components and performance through various experiments. Specifically, we focus on the impact of the patch distribution consistency loss, visualize the style editing process, assess the computational overhead of the TMPS and PRS modules, explore the effect of patch size, and compare our patch-wise approach with segmentation-based methods. This detailed examination and presentation of results not only highlight the efficacy of our Style-Editor model but also provide valuable insights into the model's operational dynamics throughout its training phase.

\subsection{The patch distribution consistency loss}
\begin{figure}[h]
    \centering
    \includegraphics[width=.7\textwidth]{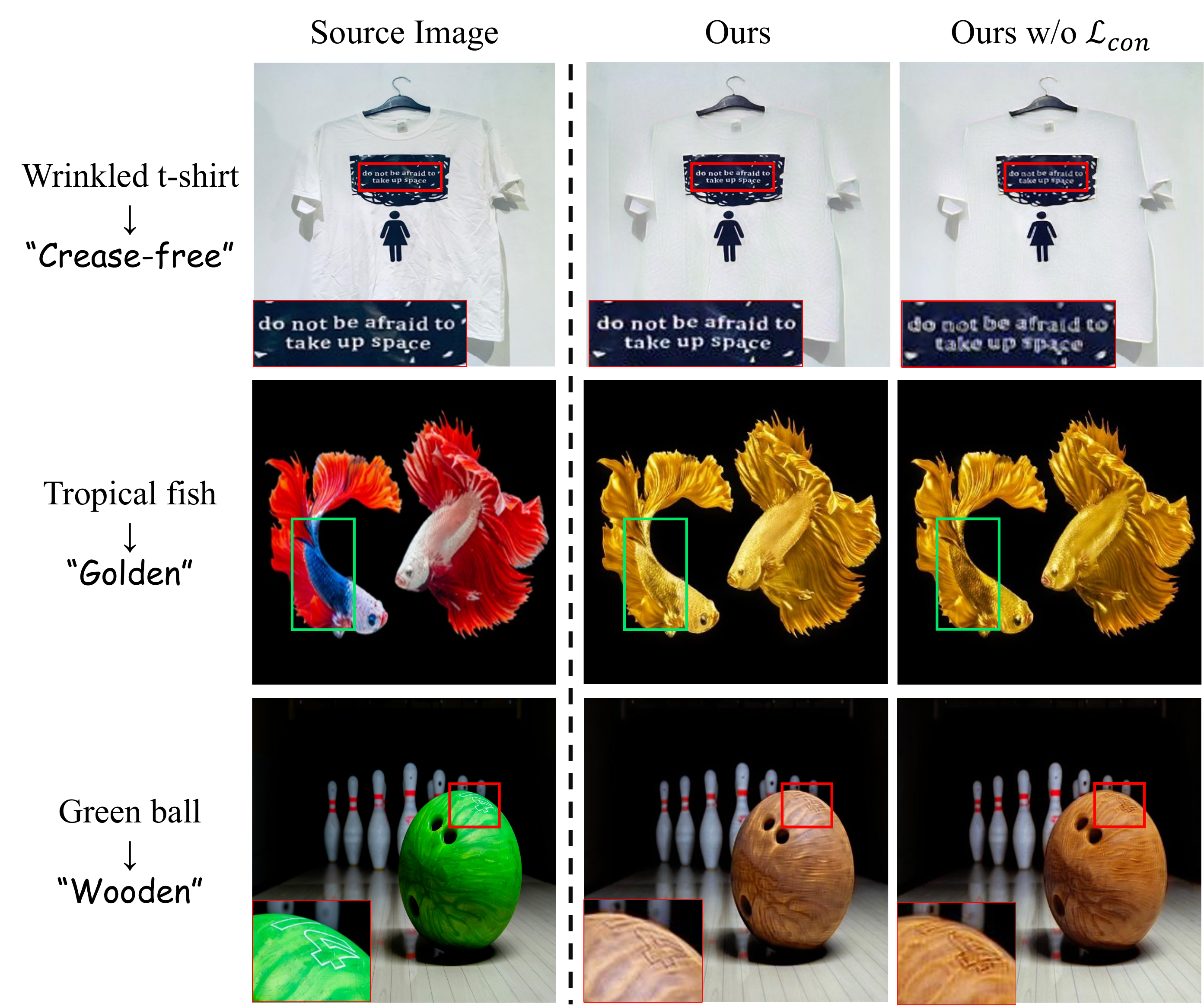} 
    \captionof{figure}{Qualitative comparison demonstrating the effect of the $\mathcal{L}_\text{pcd}$ loss by contrasting results with and without the $\mathcal{L}_\text{con}$ loss.}
    \label{fig:lcon_supp}
\end{figure}

To demonstrate the impact of the patch distribution consistency loss, denoted as $\mathcal{L}_{\text{con}}$, we conduct an ablation study. The findings from this study are displayed in \figref{fig:lcon_supp}. A key observation from our experiments is that the inclusion of $\mathcal{L}_{\text{con}}$ significantly contributes to the preservation of vital features in the images, even after the style editing process.
Notable examples include the retention of text on a T-shirt, the intricate pattern of a tropical fish, and the clarity of numbers on a bowling ball. These results highlight that the implementation of the $\mathcal{L}_{\text{con}}$ loss is crucial in maintaining a focused distribution of patches on the object. This focused approach is what enables the preservation of essential details and textures during the style editing, ensuring that the core visual elements of the object remain intact and recognizable post-transformation. This study thus underscores the effectiveness of the $\mathcal{L}_{\text{con}}$ loss in enhancing the quality and fidelity of object-centric style editings in our Style-Editor model.

\subsection{Visualization of style editing process}
\begin{figure*}[ht]
    \centering
    \includegraphics[width=.9\textwidth]{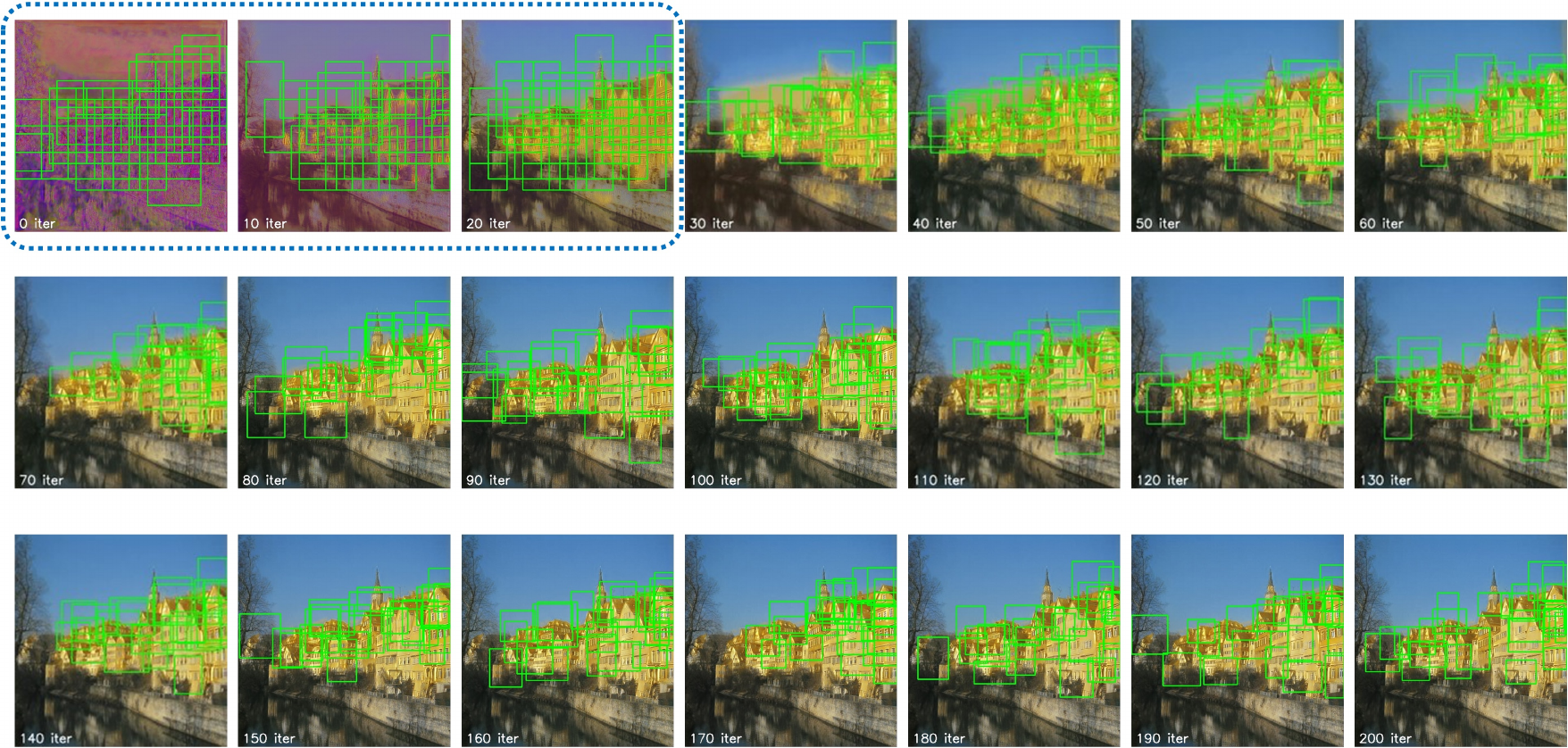} 
    \captionof{figure}{Visualization of the style editing process at intervals of every 10 iterations. This sequence illustrates the progressive transformation and refinement of the image style over time. The parts marked with blue dashed lines denote the iterations where the Pre-fixed Region Selection (PRS) module operates, and the green patches in each iteration denote the patches selected through the Text-Matched Patch Selection (TMPS). Areas not included in the patch are the regions where the Adaptive Background Preservation loss ($\mathcal{L}_{\text{abp}}$) is applied.}
    \label{fig:iter}
\end{figure*}

\figref{fig:iter} shows the iterative process of style editing as conducted by our model. Initially, the model prioritizes the preservation of the background, ensuring it remains as close to the original image as possible. This early focus on background fidelity is a crucial step in maintaining the overall integrity and context of the source image.
As the training progresses through successive iterations, the model begins to refine its approach. It gradually learns to enhance and accentuate the details within both the object and the background. This progression illustrates the model's sophisticated capability to strike a balance between maintaining background fidelity and enhancing the object of interest.
This iterative learning and adaptation process is a key aspect of our model's functionality. It demonstrates how the model evolves to effectively manage the complexities of style editing, ensuring that both the object's details and the background's essence are harmoniously preserved and enhanced.

\subsection{Computational overhead of TMPS and PRS} \label{sec:Computational overhead}
We conducted additional experiments to assess the computational overhead of the TMPS and PRS modules.
In the course of training, we processed the model with 5 distinct styles of text and 3 varieties of source images 10 times to calculate the average time spent. These time estimates encompass both the loading of the CLIP model and the image processing depicted in \tabref{tab:training_inference_times}-(a), as well as the actual training period shown in  \tabref{tab:training_inference_times}-(b). This data reveals that the proposed modules add merely an extra 10 seconds to the process (see Ours), and that the PRS can indeed reduce training time by approximately 8-9 seconds. This supports our assertion. During the inference phase, which merely involves loading a pre-saved training checkpoint and introducing an image, the presence or absence of any modules does not alter the runtime as shown in  \tabref{tab:training_inference_times}-(c), ensuring consistency across different scenarios. Thus, our experiments illustrate that while the inclusion of the TMPS and PRS modules extends training and inference times, the increase is not significantly detrimental.

\begin{table}[h]
\centering
{\fontsize{10pt}{10pt}\selectfont
\begin{tabular}{ccccc}
\toprule
\multirow{2}{*}{Method (seconds)} &
  \multirow{2}{*}{\begin{tabular}[c]{@{}c@{}}(a) Total training time\\ (+model \& data load)\end{tabular}} &
  \multirow{2}{*}{(b) Training time} &
  \multirow{2}{*}{(c) Inference time} \\
           &           &      &      \\ \midrule
w/o PRS, w/o TMPS & 40.7s & 35.2s & 0.28s \\
w/o PRS, w TMPS & 59.9s (+19.2s) & 53.3s (+18.1s) & 0.28s (+0.0s) \\
w PRS, w TMPS (Ours) & 51.9s (+11.2s) & 44.3s (+9.1s) & 0.28s (+0.0s) \\
\bottomrule
\end{tabular}}
\caption{Comparison of different methods on training and inference times}
\label{tab:training_inference_times}
\end{table}

\subsection{Impact of Patch Size}
We conducted additional experiments to investigate the impact of patch size in the TMPS module. The experimental setup and results are presented in \tabref{tab:patch_comparison} and \tabref{tab:patch_comparison_ada}, highlighting how variations in patch sizes affect key performance metrics.
Our findings demonstrate that the TMPS module remains robustness across different patch sizes. Notably, our adaptive selection of patch sizes, ranging from 64 to 128 for images of 512$\times$512 size, yields well-rounded results throughout our experiments. This adaptive approach reflects a careful trade-off: smaller patches enhance image quality metrics, such as PSNR, $\text{Con}_F$, and $\text{Con}_B$, while larger patches help preserve stylization consistency, especially around object boundaries. Nevertheless, it is worth noting that such a reduction in patch size might also decrease the CLIP similarity for the foreground ($\text{Sim}_F$), potentially compromising the content's integrity. 

\begin{table*}[ht]
\centering
{
\begin{tabular}
{c@{\hspace{1mm}}|c@{\hspace{1mm}}c@{\hspace{1mm}}|c@{\hspace{1mm}}c@{\hspace{1mm}}c@{\hspace{1mm}}c@{\hspace{1mm}}c@{\hspace{1mm}}c@{\hspace{1mm}}c@{\hspace{1mm}}c@{\hspace{1mm}}c@{\hspace{1mm}}c@{\hspace{1mm}}}
\toprule
   & \multicolumn{2}{c|}{Foreground quality metrics} &\multicolumn{6}{c}{Background quality metrics}\\
  Methods  & \multicolumn{1}{c}{\hspace{4mm} $\text{Sim}_{F}\uparrow$} & \multicolumn{1}{c|}{$\text{Con}_{F}\downarrow$} & \multicolumn{1}{c}{$\text{L1}_{B}\downarrow$} & \multicolumn{1}{c}{$\text{Con}_{B}\downarrow$} & \multicolumn{1}{c}{$\text{Sty}_{B}\downarrow$} & \multicolumn{1}{c}{$\text{SSIM}_{B}\uparrow$} & \multicolumn{1}{c}{$\text{DISTS}_{B}\downarrow$} & \multicolumn{1}{c}{$\text{PSNR}_{B}\uparrow$} \\
   
\midrule
patch 64 & 0.3 & \textbf{3.46} & \textbf{0.09} & \underline{1.2} & \underline{0.11} & \textbf{0.90} & \underline{0.08} & \textbf{27.92} \\
patch 96 & \underline{0.32} & 3.94 & \textbf{0.09} & 1.25 & 0.12 & \underline{0.89} & \underline{0.08} & 27.56 \\
patch 128 & \textbf{0.33} & 4.24 & \underline{0.10} & 1.29 & 0.13 & \underline{0.89} & \underline{0.08} & 27.17 \\
\midrule
\textbf{Ours (patch 64-128)} & \textbf{0.33} & \underline{3.75} & \underline{0.10} & \textbf{1.15} & \textbf{0.10} & \textbf{0.90} & \textbf{0.07} & \underline{27.65} \\
\bottomrule

\end{tabular}
}
\caption{Quantitative evaluation of fixed patch sizes. Metrics include similarity (Sim$_F$, L1$_B$), content loss (Con$_F$, Con$_B$), style loss (Sty$_B$), structural similarity index (SSIM), perceptual distance (DISTS), and peak signal-to-noise ratio (PSNR). $\uparrow$ indicates higher values are better, while $\downarrow$ indicates lower values are better.}
\label{tab:patch_comparison}
\end{table*}

\begin{table*}[ht]
\centering
{
\begin{tabular}
{c@{\hspace{1mm}}|c@{\hspace{1mm}}c@{\hspace{1mm}}|c@{\hspace{1mm}}c@{\hspace{1mm}}c@{\hspace{1mm}}c@{\hspace{1mm}}c@{\hspace{1mm}}c@{\hspace{1mm}}c@{\hspace{1mm}}c@{\hspace{1mm}}c@{\hspace{1mm}}c@{\hspace{1mm}}}
\toprule
   & \multicolumn{2}{c|}{Foreground quality metrics} &\multicolumn{6}{c}{Background quality metrics}\\
  Methods  & \multicolumn{1}{c}{\hspace{4mm} $\text{Sim}_{F}\uparrow$} & \multicolumn{1}{c|}{$\text{Con}_{F}\downarrow$} & \multicolumn{1}{c}{$\text{L1}_{B}\downarrow$} & \multicolumn{1}{c}{$\text{Con}_{B}\downarrow$} & \multicolumn{1}{c}{$\text{Sty}_{B}\downarrow$} & \multicolumn{1}{c}{$\text{SSIM}_{B}\uparrow$} & \multicolumn{1}{c}{$\text{DISTS}_{B}\downarrow$} & \multicolumn{1}{c}{$\text{PSNR}_{B}\uparrow$} \\
   
\midrule
patch 32-64 & 0.29 & \textbf{2.90} & \textbf{0.09} & \textbf{1.12} & \underline{0.11} & \textbf{0.91} & \underline{0.08} & \textbf{28.41} \\
\textbf{Ours (patch 64-128)} & \underline{0.33} & \underline{3.75} & \underline{0.10} & \underline{1.15} & \textbf{0.10} & \underline{0.90} & \textbf{0.07} & \underline{27.65} \\
patch 128-256 & \textbf{0.34} & 4.43 & 0.11 & 1.37 & 0.15 & 0.87 & 0.09 & 26.05 \\
\bottomrule
\end{tabular}
}
\caption{Quantitative evaluation of adaptive patch sizes. Metrics include similarity (Sim$_F$, L1$_B$), content loss (Con$_F$, Con$_B$), style loss (Sty$_B$), structural similarity index (SSIM), perceptual distance (DISTS), and peak signal-to-noise ratio (PSNR). $\uparrow$ indicates higher values are better, while $\downarrow$ indicates lower values are better.}
\label{tab:patch_comparison_ada}
\end{table*}

\subsection{Comparison with Segmentation method}
\begin{figure*}[h]
    \centering
\includegraphics[width=.9\textwidth]{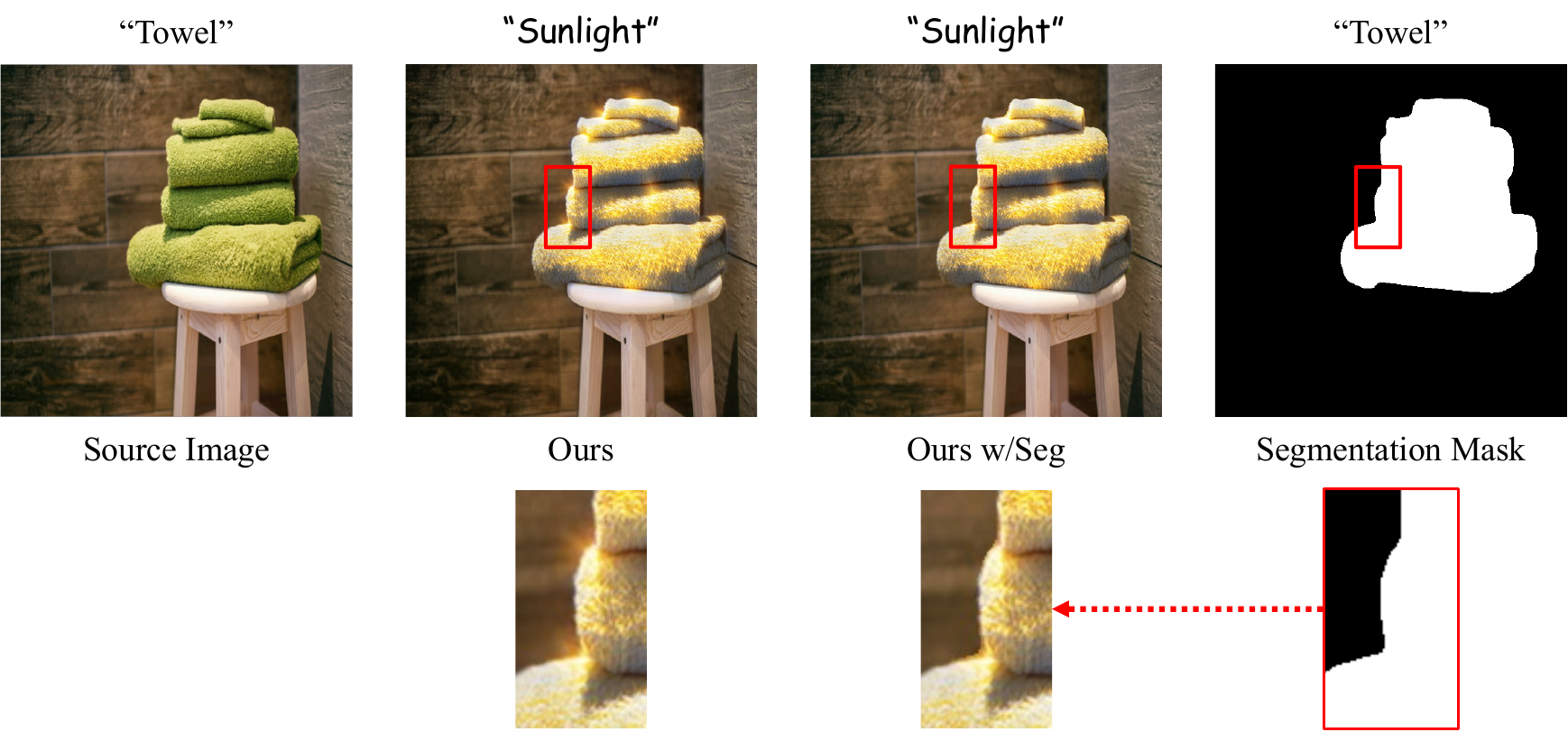} 
    \captionof{figure}{Qualitative comparison between our results and our results with segmentation masks applied.}
    \label{fig:vs_seg}
\end{figure*}

To evaluate the effectiveness of our approach compared to segmentation-based methods, we conducted experiments by applying a mask to our results using the open-vocabulary segmentation model ODISE \cite{xu2023open}. The comparative results are illustrated in \figref{fig:vs_seg}.
When applying a `Sunlight' style to a towel image, our model generates results where light naturally diffuses around the object.
In contrast, segmentation-based methods rely on per-pixel binary classification (0 or 1) to determine the presence of an object, resulting in abrupt and unnatural transitions at the boundaries where the style is applied. This disrupts the overall continuity and realism of the stylized image.
Furthermore, segmentation-based methods typically require additional training on datasets, such as the MSCOCO \cite{lin2014microsoft} dataset.
These findings highlight the advantage of our model in maintaining seamless and natural style editing, avoiding the artifacts commonly observed at object boundaries in segmentation-based methods. Moreover, by leveraging the CLIP encoder for both style transfer and object patch identification, our approach eliminates the need for pre-trained segmentation networks, achieving both simplicity and effectiveness.

\section{Visualization of evaluation metric}
\begin{figure}[h]
    \centering
    \includegraphics[width=.85\textwidth]{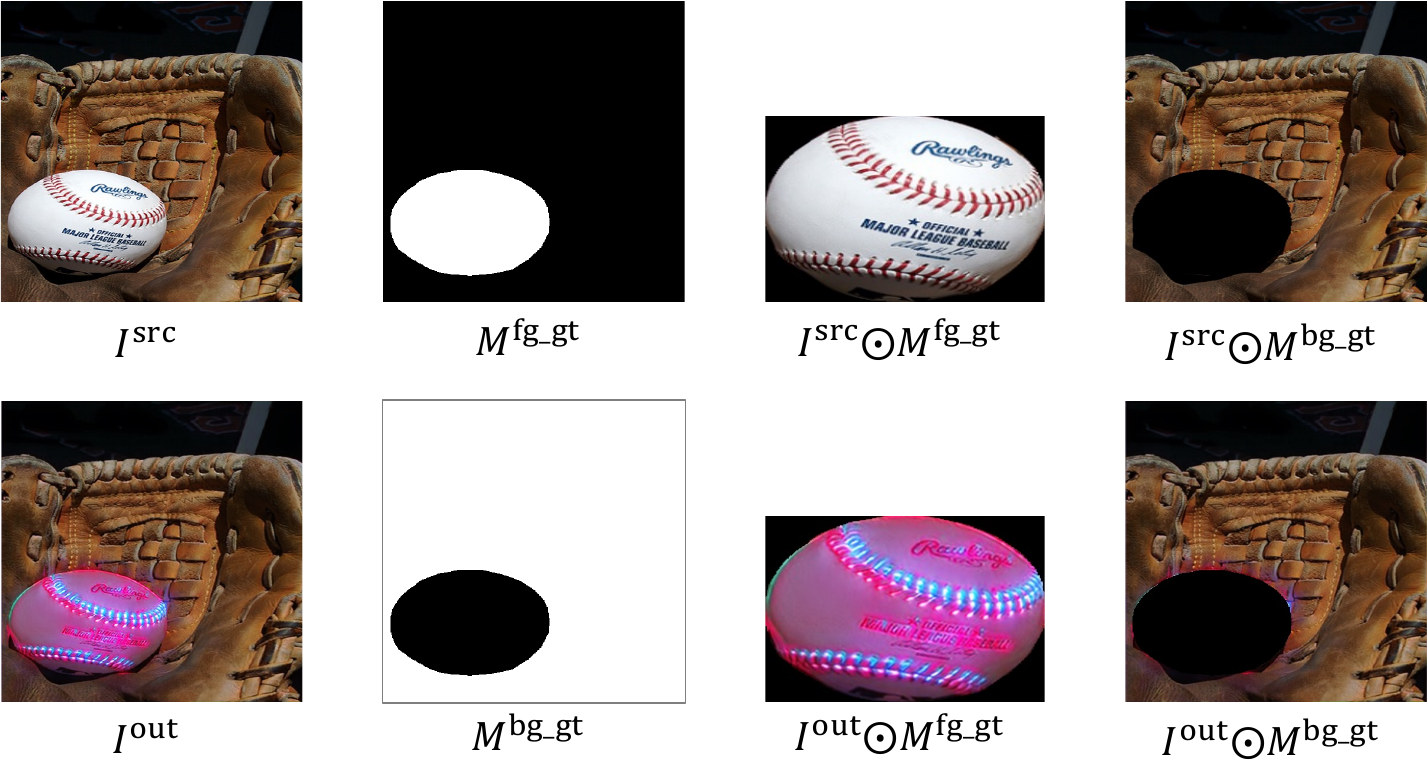} 
    \captionof{figure}{Example images for evaluating foreground quality metrics and background quality metrics.}
    \label{fig:eval_vis}
\end{figure}

To evaluate the performance of object-centric style editing, we conducted evaluations using the annotations from the MS COCO 2017 dataset.
\figref{fig:eval_vis} presents the visualized results used as reference examples for the evaluation metrics employed in our experiments.
In particular, $M^{\text{fg\_gt}}$ denotes the ground truth (GT) mask corresponding to the target class.
Using this mask, we calculated the foreground quality metrics by masking regions outside the class object areas and cropping the images according to the mask area, as shown in $I^{\text{src}}\odot M^{\text{fg\_gt}}$ and $I^{\text{out}}\odot M^{\text{fg\_gt}}$ in the \figref{fig:eval_vis}.
Conversely, the background quality metrics were measured using images where the object regions were excluded, as represented by $I^{\text{src}}\odot M^{\text{bg\_gt}}$ and $I^{\text{out}}\odot M^{\text{bg\_gt}}$ in the \figref{fig:eval_vis}.

\clearpage
\section{Complex input texts}
\begin{figure*}[h]
    \centering
    \includegraphics[width=.85\textwidth]{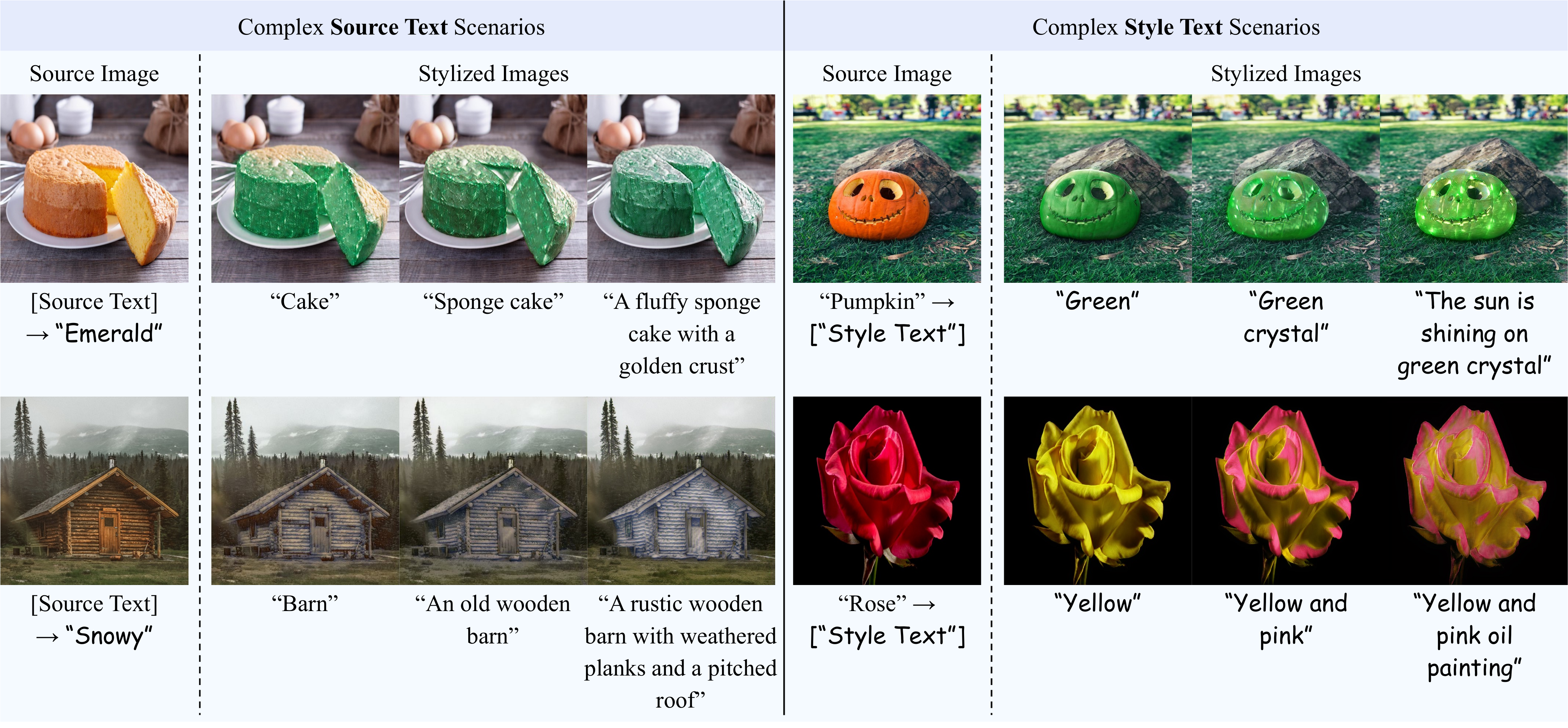} 
    \captionof{figure}{Our style editing results with complex image-text scenarios.}
    \label{fig:complex}
\end{figure*}

We have conducted further experiments using a diverse set of examples featuring intricate ‘source’ and ‘style’ texts resulting in \figref{fig:complex}.  For example, in the ‘cake to emerald’ scenario, although the stylized image with simple text retains aspects of the original cake's style, the detailed source text enables the creation of a cake styled purely in emerald. Similarly, in the ‘barn to snowy’ scenario, the model adeptly preserves the background's style while effectively applying style editing to the foreground object. Furthermore, our experiments incorporating complex ‘style’ texts illustrate the complexity of the text does not impede our method's ability to achieve the intended editing effects. These results clearly demonstrate that our method is adept not only at managing simple texts but is equally effective with complex textual inputs. This expansion of our testing framework underscores our method's robustness and versatility in accommodating a broad spectrum of textual complexities.

\section{Limitations of Style-Editor}
\begin{figure}[h]
    \centering
    \includegraphics[width=.55\textwidth]{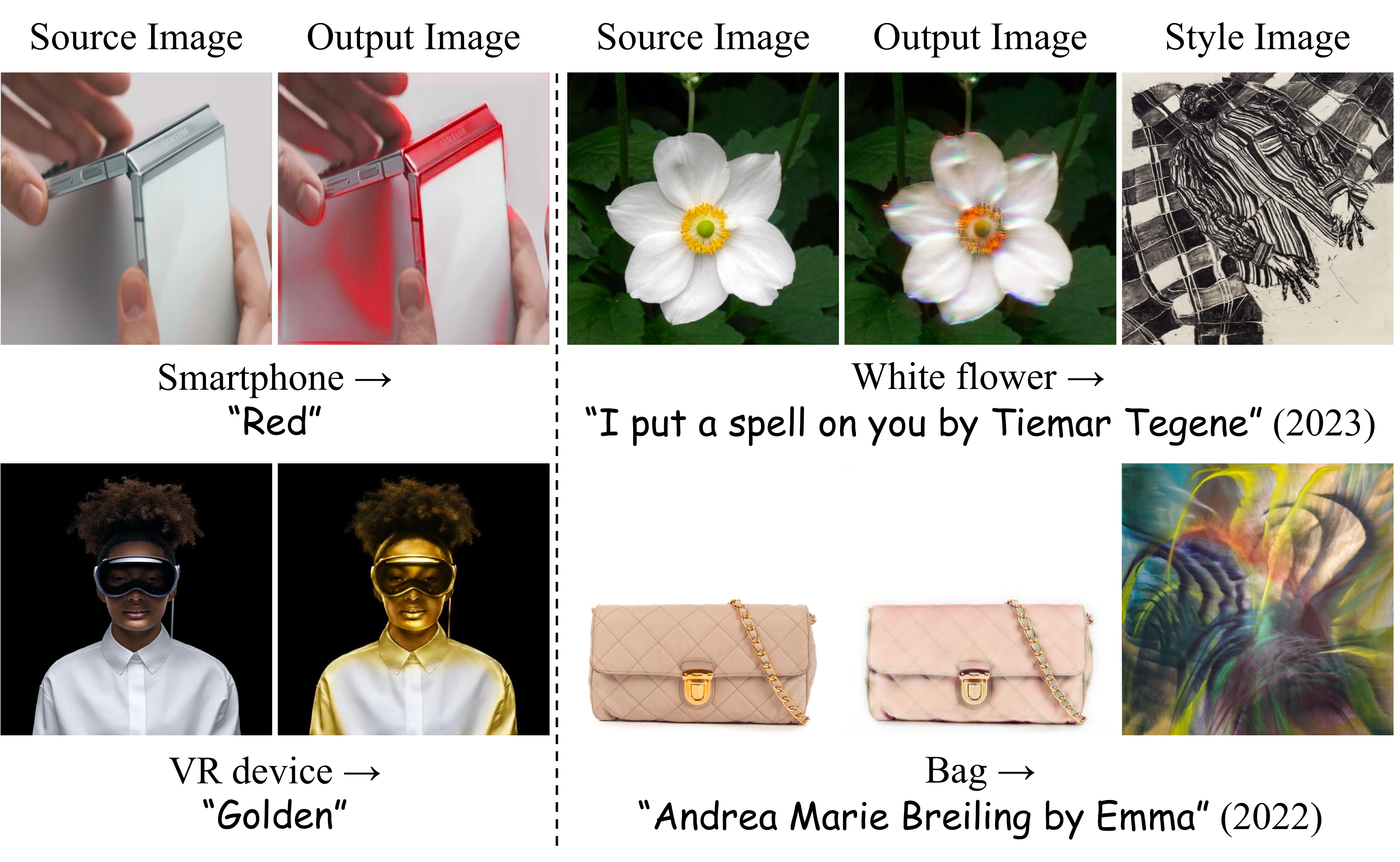} 
    \captionof{figure}{Failure cases of Style-Editor. The numbers following the style text denote the year of the art work was created. }
    \label{fig:failure}
\end{figure}

A noted limitation of Style-Editor is its reliance on the CLIP model's feature space and classification capabilities. Consequently, its performance may diminish for styles or objects less represented or absent in the CLIP training dataset, such as recent art or gadgets, as illustrated in \figref{fig:failure}. This dependency highlights a potential area for future development, aiming to enhance Style-Editor's adaptability to emerging styles and objects.

\clearpage
\section{User study}
\begin{figure*}[ht]
    \centering
\includegraphics[width=.9\textwidth]{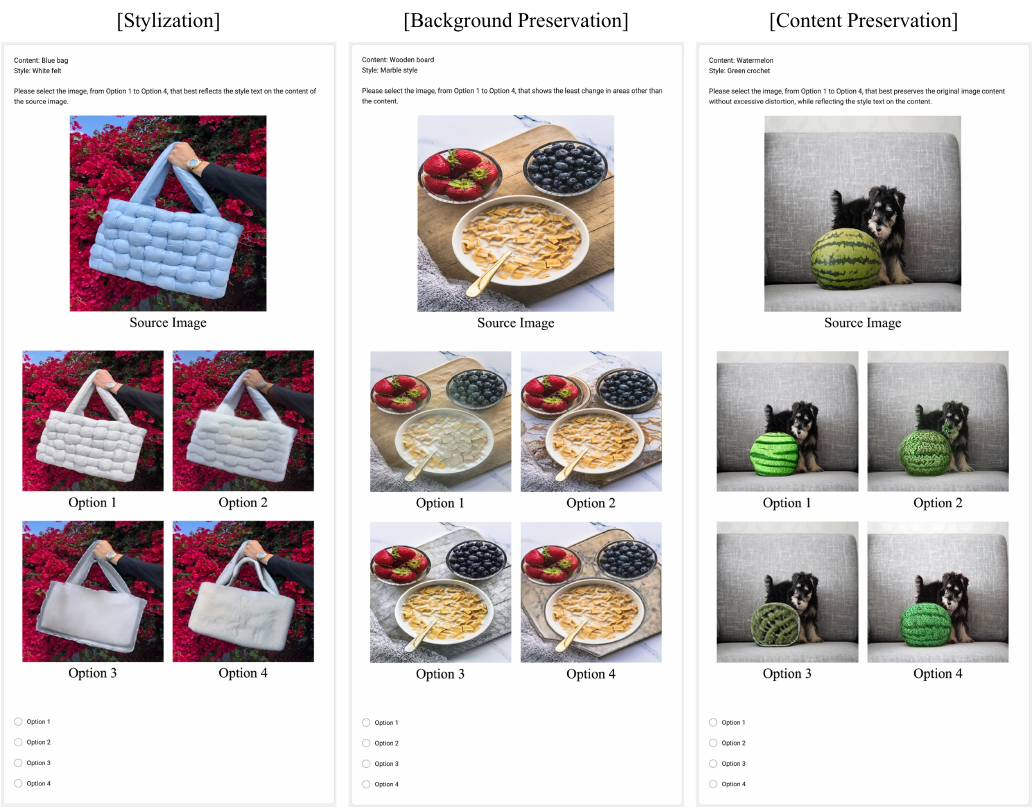} 
    \captionof{figure}{User study question examples. The order of options was shuffled for each question.}
    \label{fig:user_question}
    
\end{figure*}

We conducted two rounds of user studies to enhance our evaluation process, the results of which are presented in \tabref{tab:user_study}. This study involved 50 participants, whose ages ranged from their 20s to 50s. We designed the study to assess three key aspects: stylization quality, background preservation, and content preservation, providing nine examples within each category for evaluation. The evaluation included the mask-based editing models Glide \cite{nichol2021glide}, and Blended Diffusion \cite{avrahami2022blended}, along with the top-4 models from \tabref{tab:vs_text} --- FlexIT \cite{couairon2022flexit}, LEDITS ++ \cite{brack2024ledits++}, Null text inversion \cite{mokady2023null}, and Text2LIVE~\cite{bar2022text2live}. In each round, we compared the stylized images produced by three competing models with those generated by our model. Our analysis indicates that our model excels in achieving a balanced object-centric style editing, effectively maintaining the semantic integrity of the object and the background without compromise. In particular, compared with mask-based methods, our model delivers more natural and stable performance. The user study question examples are illustrated in \figref{fig:user_question}.

\begin{table}[h]
\centering
\begin{tabular}{c|ccc}
\toprule
    Method&  StyQual $\uparrow$ & BackPre $\uparrow$ & ContPre $\uparrow$ \\ 
    \midrule
    Blended Diffusion \cite{avrahami2022blended}       & 4.4 \% & 7.3 \% & 2.4 \% \\
    Glide \cite{nichol2021glide}   & 1.1 \% & 12.0 \% & 3.6 \%\\
    Text2LIVE \cite{bar2022text2live}  & 29.3 \% & 14.9 \% & 11.8 \%\\
    \midrule
    FlexIT \cite{couairon2022flexit}       & 14 \% & 5.1 \% & 2.9 \% \\
    LEDITS++ \cite{brack2024ledits++}   & 13.3 \% & 6 \% & 6.2 \%\\
    Null-text inversion \cite{mokady2023null}  & 5.8 \% & 25.8 \% & 16 \%\\
    \midrule
    Ours (Avg.) & \textbf{66} \% & \textbf{64.4} \% & \textbf{78.6} \%\\ 
    \bottomrule
\end{tabular}
\caption{User study detailing preference percentages.  }
\label{tab:user_study}
\end{table}

\section{Additional quantitative results}
To assess the robustness of our model, we conducted additional quantitative experiments on the top-5 models from \tabref{tab:vs_text}, with training time evaluated following the settings in \secref{sec:Computational overhead}. The test set was expanded by randomly selecting 50 images from the MSCOCO 2017 dataset and pairing each with 10 different style text descriptions, yielding a total of 500 stylized images for evaluation.
As presented in \tabref{tab:vs_text_addi}, our model consistently outperforms others on the extended test set while ranking second in speed. Notably, even compared to LEDITS++ \cite{brack2024ledits++}, the fastest model, our approach demonstrates superior performance in both foreground and background metrics. These results validate its strong generalization ability and highlight its well-balanced stylized output at a relatively fast pace.

\begin{table*}[ht]
\centering
{
\begin{tabular}
{c@{\hspace{1mm}}|c@{\hspace{1mm}}c@{\hspace{1mm}}|c@{\hspace{1mm}}c@{\hspace{1mm}}c@{\hspace{1mm}}c@{\hspace{1mm}}c@{\hspace{1mm}}c@{\hspace{1mm}}|c@{\hspace{1mm}}c@{\hspace{1mm}}}
\toprule
   & \multicolumn{2}{c|}{Foreground metrics} &\multicolumn{6}{c|}{Background metrics} & \multicolumn{1}{c}{} \\
  Methods  & \multicolumn{1}{c}{$\text{Sim}_{F}\uparrow$} & \multicolumn{1}{c|}{$\text{Con}_{F}\downarrow$} & \multicolumn{1}{c}{$\text{L1}_{B}\downarrow$} & \multicolumn{1}{c}{$\text{Con}_{B}\downarrow$} & \multicolumn{1}{c}{$\text{Sty}_{B}\downarrow$} & \multicolumn{1}{c}{$\text{SSIM}_{B}\uparrow$} & \multicolumn{1}{c}{$\text{DISTS}_{B}\downarrow$} & \multicolumn{1}{c|}{$\text{PSNR}_{B}\uparrow$} & \multicolumn{1}{c}{Time (s) $\downarrow$} \\
\midrule
FlexIT~\cite{couairon2022flexit} & 0.24 & 7.03 & 0.20 & 4.09 & 0.36 & 0.66 & 0.14 & 21.17 & 59.8 \\
LEDITS++~\cite{brack2024ledits++} & 0.21 & 5.98 & 0.19 & 2.70 & 0.42 & 0.75 & 0.13 & 21.43 & \textbf{10.4} \\
NTI~\cite{mokady2023null} & 0.20 & 4.64 & 0.16 & 3.02 & 0.32 & 0.74 & 0.12 & 23.46 & 102.3 \\
LPM~\cite{patashnik2023localizing} & 0.20 & 8.62 & 0.25 & 4.81 & 0.73 & 0.67 & 0.19 & 19.53 & 119.9 \\
Text2LIVE~\cite{bar2022text2live} & \underline{0.30} & \underline{3.65} & \underline{0.13} & \underline{1.25} & \underline{0.17} & \underline{0.87} & \textbf{0.08} & \underline{25.47} & 412.6 \\
\midrule
\textbf{Ours} & \textbf{0.31} & \textbf{3.09} & \textbf{0.10} & \textbf{0.95} & \textbf{0.08} & \textbf{0.89} & \textbf{0.08} & \textbf{26.71} & \underline{44.3} \\
\bottomrule

\end{tabular}
}
\caption{Additional quantitative comparison with 500 stylized images. $\uparrow$ indicates that higher values are better, while $\downarrow$ indicates that lower values are better.}
\label{tab:vs_text_addi}
\end{table*}

\clearpage
\section{Additional object-centric style editing results using our Style-Editor }
\begin{figure*}[h]
    \centering
    \includegraphics[width=1\textwidth]{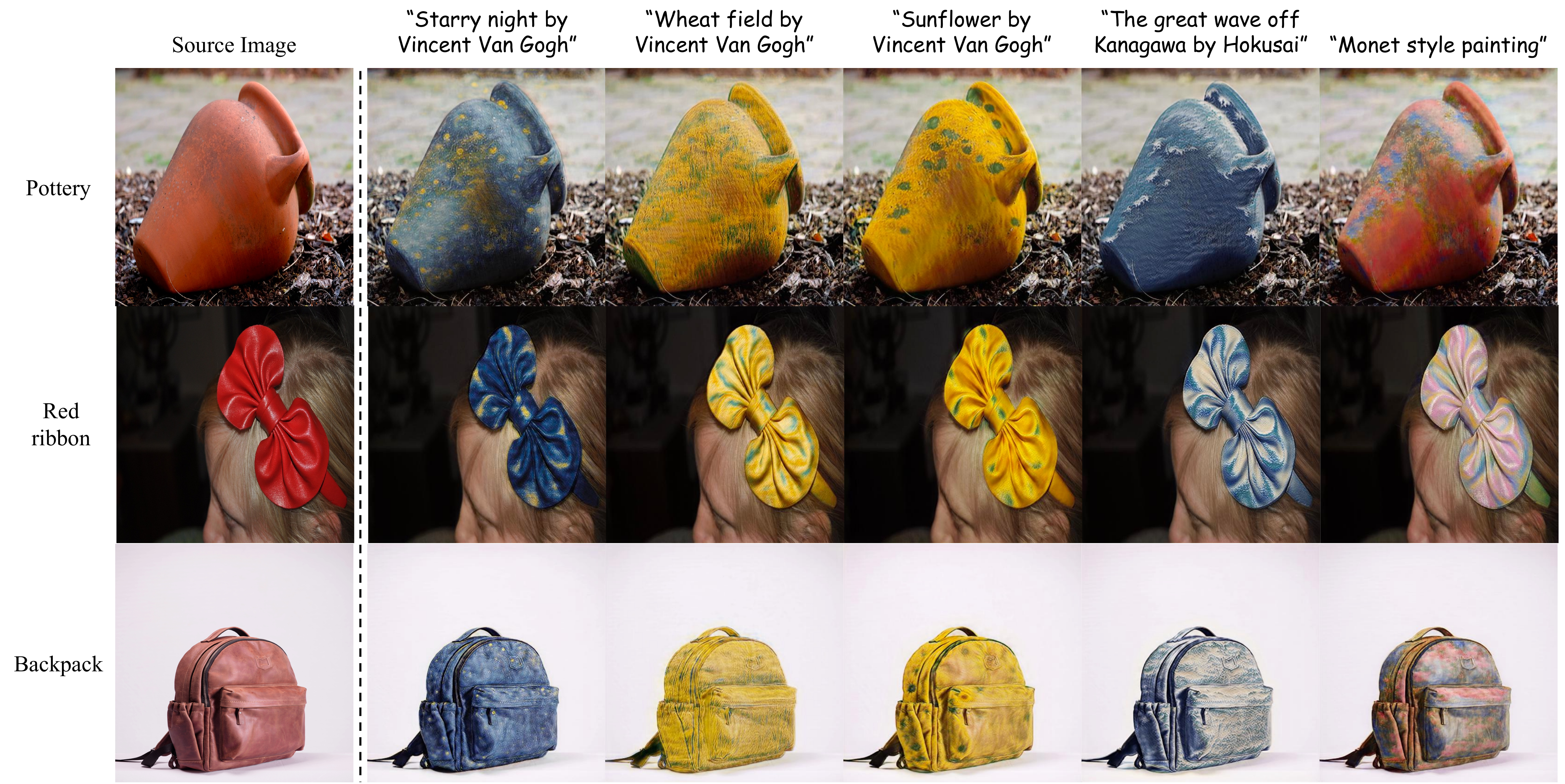} 
    \captionof{figure}{Stylization results demonstrating various \textbf{``artistic"} styles guided by text using our Style-Editor model.}
    \label{fig:painting}
\end{figure*}

\begin{figure}[h]
    \centering
    \includegraphics[width=.9\textwidth]{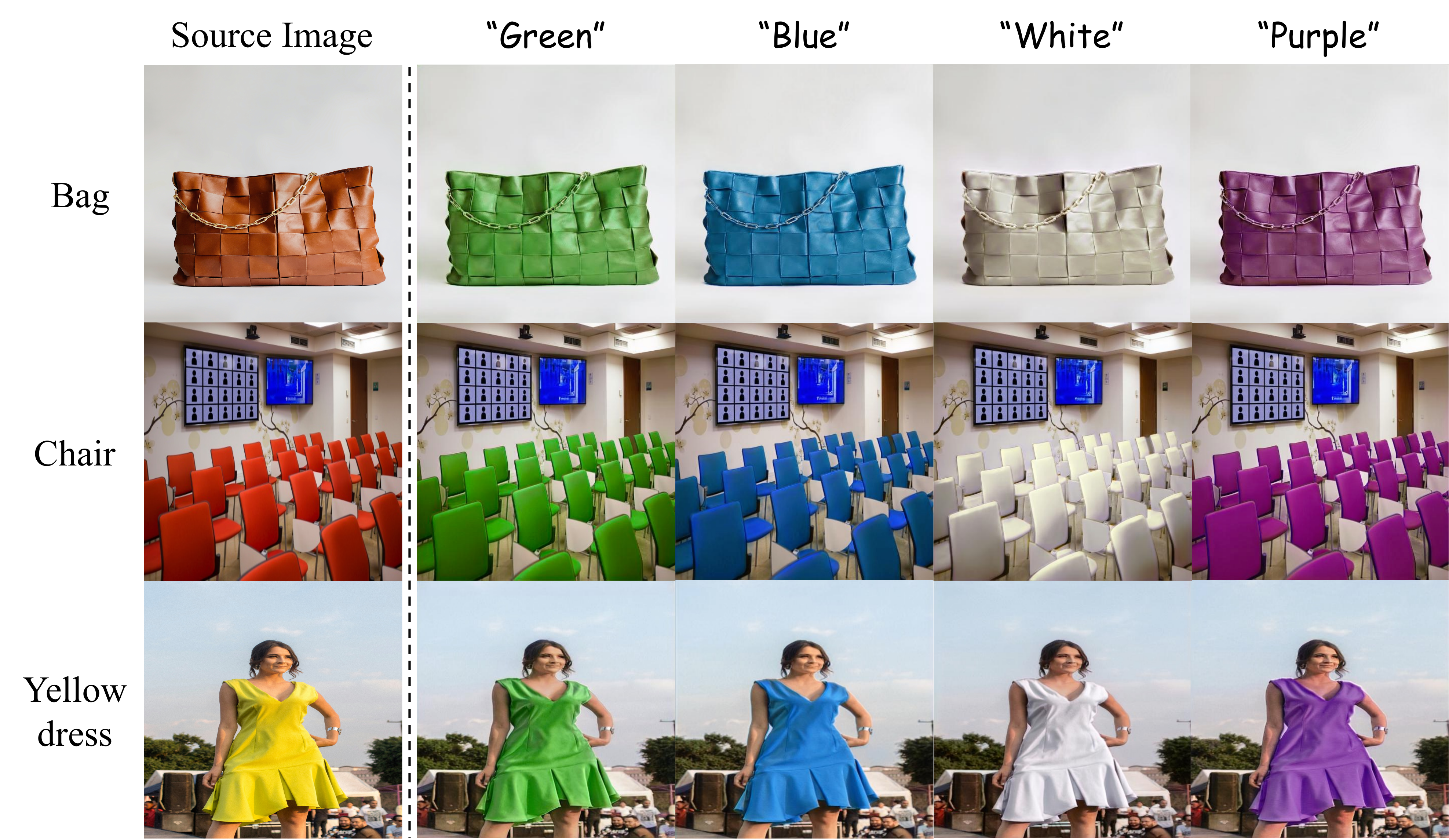} 
    \captionof{figure}{Stylization results demonstrating various \textbf{``color"} styles guided by text using our Style-Editor model.}
    \label{fig:color}
\end{figure}

\begin{figure}[h]
    \centering
    \includegraphics[width=.9\textwidth]{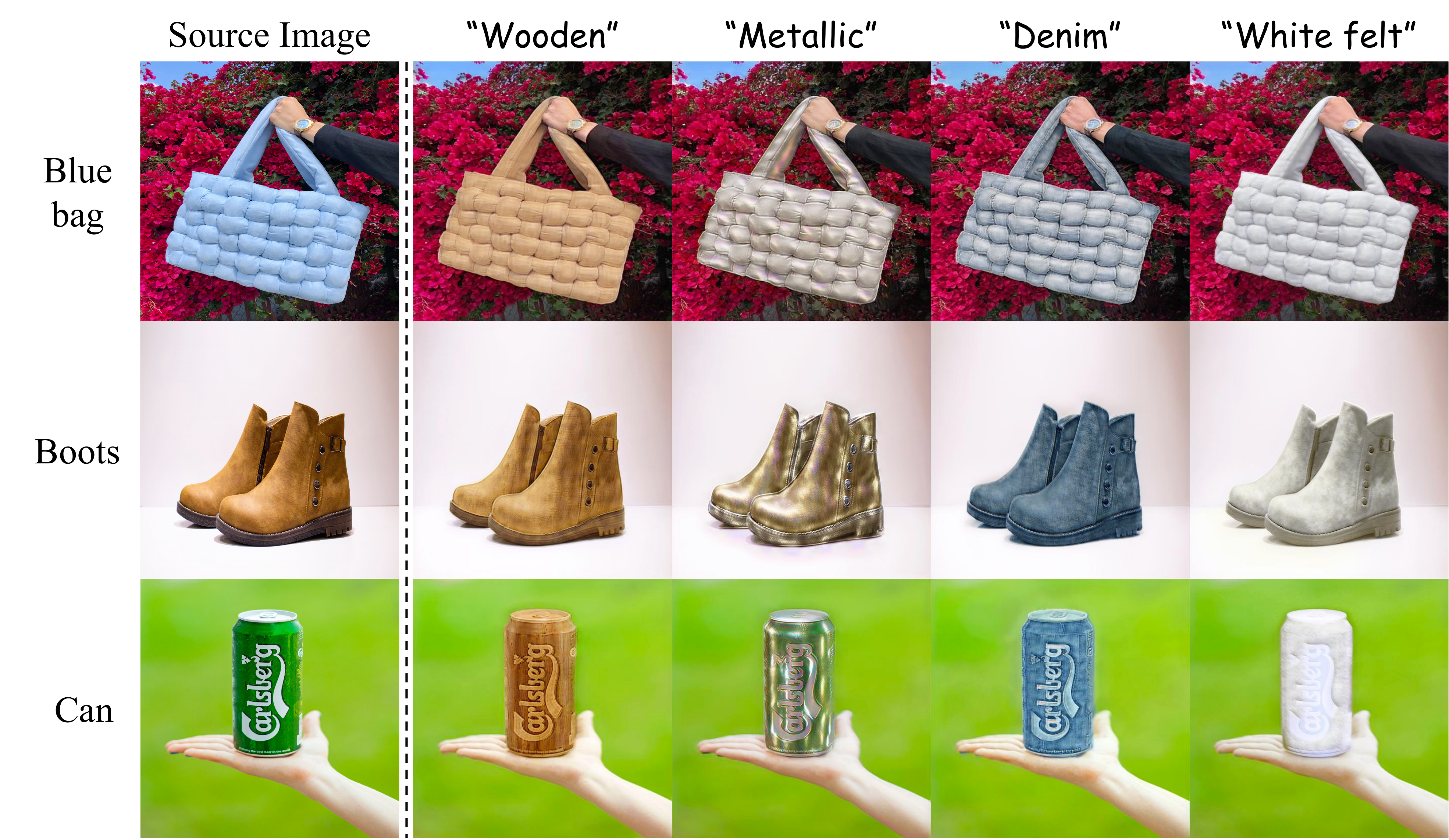} 
    \captionof{figure}{Stylization results demonstrating various \textbf{``texture"} styles guided by text using our Style-Editor model.}
    \label{fig:texture}
\end{figure}

\begin{figure*}[h]
    \centering
    \includegraphics[width=.86\textwidth]{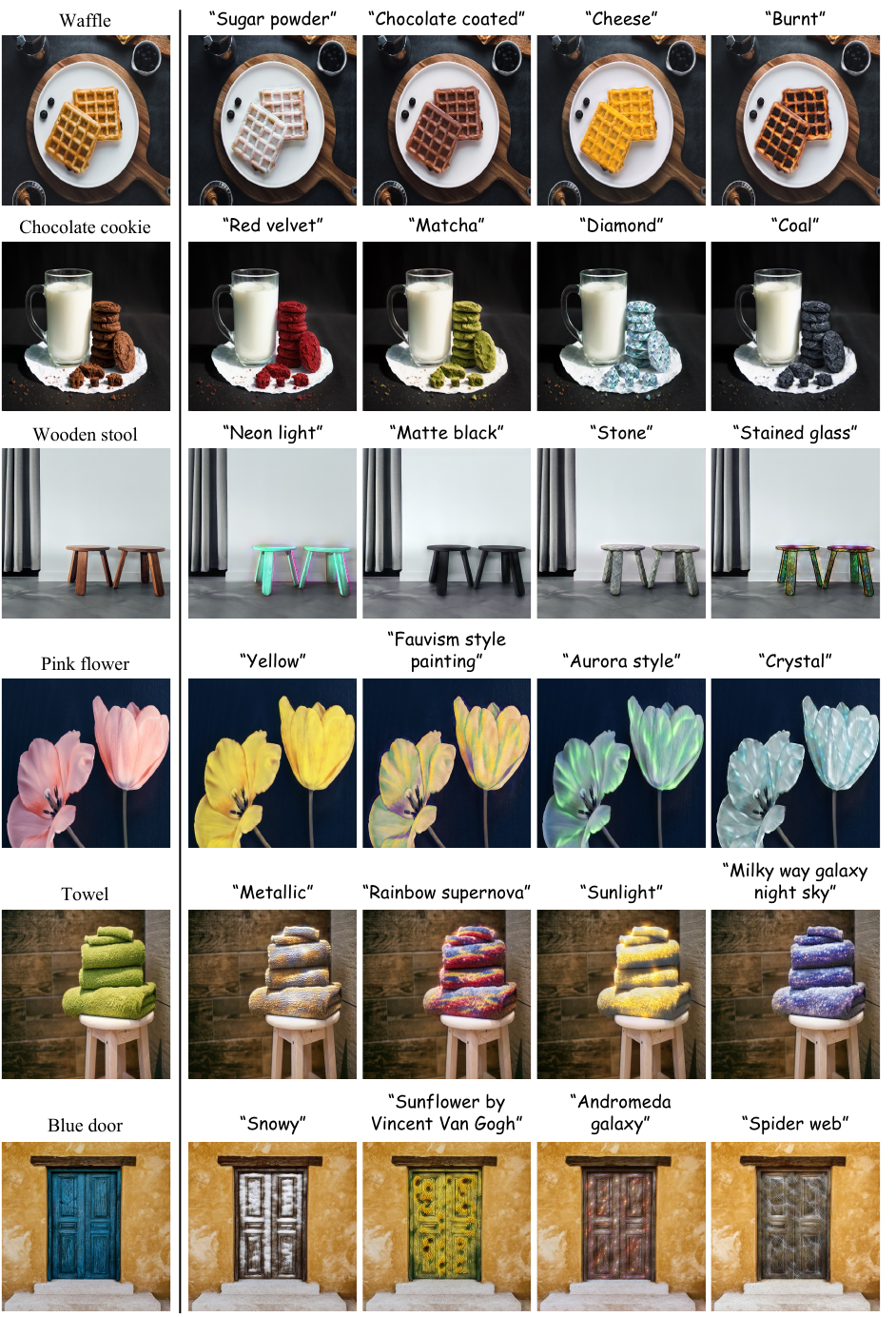} 
    \captionof{figure}{Our additional stylization results with various image-text scenarios.}
    \label{fig:figonly}
\end{figure*}

We show additional results of our Style-Editor method across diverse scenes, as seen in \figref{fig:painting}, \figref{fig:color}, \figref{fig:texture}, and \figref{fig:figonly}. These figures illustrate the effectiveness of our approach in various style editing scenarios, emphasizing the versatility of Style-Editor.
TMPS plays a crucial role in initiating the style editing process. It specifically targets image areas that correspond with the input text, ensuring that the chosen style is applied seamlessly and appropriately to the relevant objects. This targeted approach results in a harmonious blend of the new style with the original image, particularly in areas corresponding to the source text.
Furthermore, our method incorporates an innovative ABP loss. This component of Style-Editor is vital in maintaining the integrity of the background areas during the style editing process. It ensures that these areas remain unaffected by the changes applied to the object of interest. This loss function is key to achieving a balanced and natural-looking result where the style changes are confined to the targeted object, while the rest of the image retains its original appearance.
The results in \figref{fig:painting}, \figref{fig:color},  \figref{fig:texture}, and \figref{fig:figonly} collectively showcase the robust customizability and adaptability of the Style-Editor method. They demonstrate its capability to handle a diverse range of styles and scenarios, effectively adapting the chosen style to the specific objects in the image as dictated by the source text, all while preserving the overall aesthetic and integrity of the background. 

\vspace{20pt}

\end{document}